\documentclass[10pt,twocolumn,letterpaper]{article}

\usepackage{wacv}
\usepackage{times}
\usepackage{epsfig}
\usepackage{graphicx}
\usepackage{amsmath}
\usepackage{amssymb}

\usepackage[font=small]{caption} 
\usepackage[hyphens]{url}  % DO NOT CHANGE THIS
\usepackage{graphicx} % DO NOT CHANGE THIS
\usepackage{graphicx}  % DO NOT CHANGE THIS
\usepackage{nccmath}
\usepackage{enumitem}
\usepackage{color}

\def\wacvPaperID{1289} % Enter the WACV Paper ID here

\wacvfinalcopy % *** Uncomment this line for the final submission
\ifwacvfinal
\fi

\ifwacvfinal
\usepackage[breaklinks=true,bookmarks=false]{hyperref}
\else
\usepackage[pagebackref=true,breaklinks=true,colorlinks,bookmarks=false]{hyperref}
\fi

\ifwacvfinal
\else
\fi

\begin{document}

%%%%%%%%% TITLE
\title{Conditional Link Prediction of Category-Implicit Keypoint Detection}

\author{Ellen Yi-Ge$^1$, Rui Fan$^2$, Zechun Liu$^1$, Zhiqiang Shen$^1$\\
$^1$ Carnegie Mellon University\\
$^2$ UC San Diego\\
}

\maketitle
%\thispagestyle{empty}

%%%%%%%%% ABSTRACT
\begin{abstract}
Keypoints of objects reflect their concise abstractions, while the corresponding connection links (CL) build the skeleton by detecting the intrinsic relations between keypoints. Existing approaches are typically computationally-intensive, inapplicable for instances belonging to multiple classes, and/or infeasible to simultaneously encode connection information. To address the aforementioned issues, we propose an end-to-end category-implicit Keypoint and Link Prediction Network (KLPNet), which is the first approach for simultaneous semantic keypoint detection (for multi-class instances) and CL rejuvenation.
In our KLPNet, a novel Conditional Link Prediction Graph is proposed for link prediction among keypoints that are contingent on a predefined category. Furthermore, a Cross-stage Keypoint Localization Module (CKLM) is introduced to explore feature aggregation for coarse-to-fine keypoint localization. Comprehensive experiments conducted on three publicly available benchmarks demonstrate that our KLPNet consistently outperforms all other state-of-the-art approaches. Furthermore, the experimental results of CL prediction also show the effectiveness of our KLPNet with respect to occlusion problems.
\end{abstract}
\begin{figure}[t!]
\centering
\includegraphics[width=\linewidth]{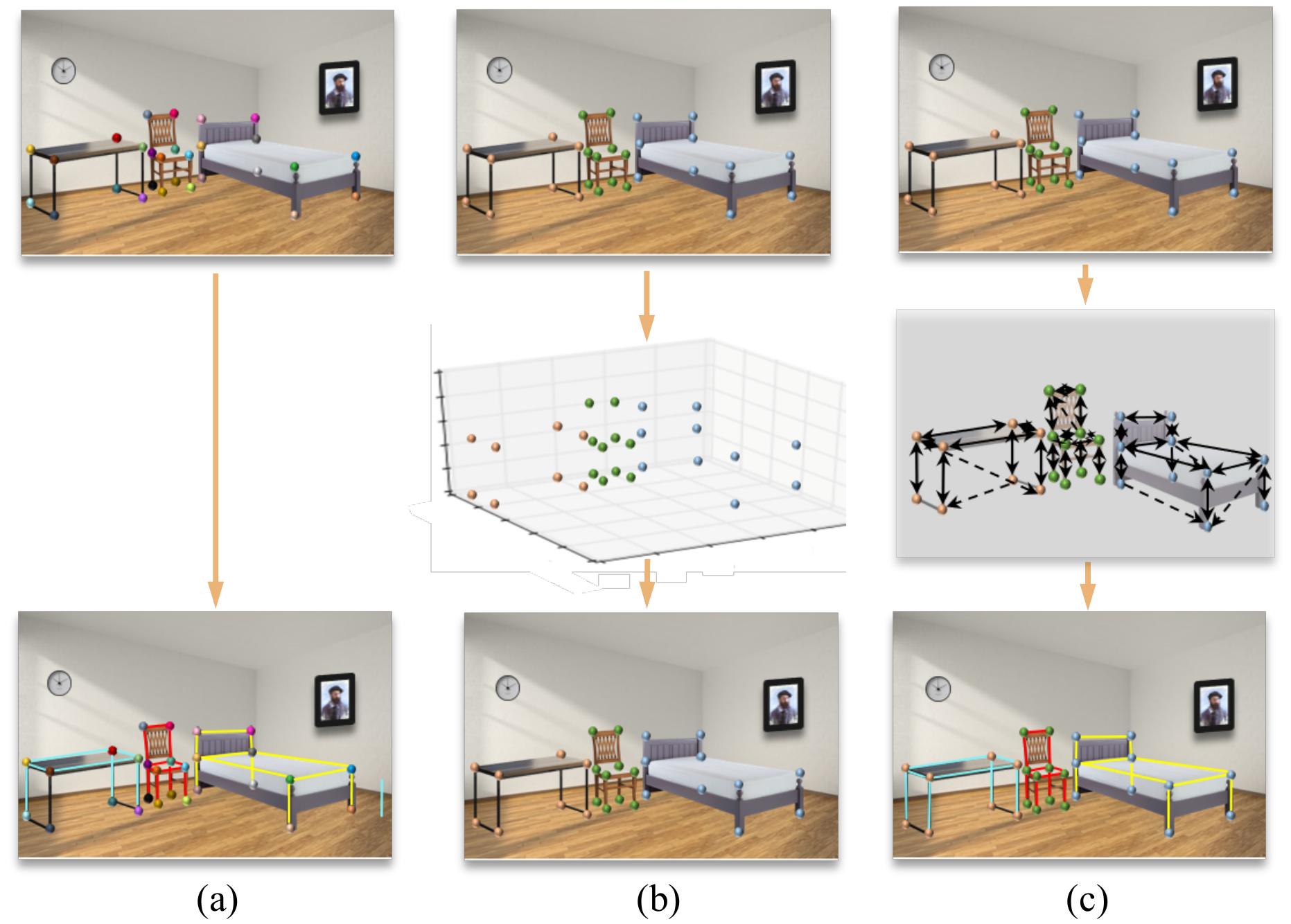}
\caption{(a) KeypointNet \cite{suwajanakorn2018discovery}. (b) StarMap \cite{zhou2018starmap}. (c) Our proposed Keypoint and Link Prediction Network (KLPNet), a category-implicit approach. KLPNet is the first framework capable of finding connection link for multi-class object keypoints. KLPNet adopts a conditional embedding graph to implement the link prediction based on the extracted features from the 2D images.}
\vspace{-1.2em}
\label{fig:approach}
\end{figure}

%%%%%%%%% BODY TEXT
\section{Introduction}
Accurate semantic keypoint localization and detection is the basic prerequisite for copious computer vision applications, including simultaneous localization and mapping \cite{sualeh2019simultaneous}, human pose estimation \cite{li2019rethinking}, hand key-joint estimation \cite{pang2020automatic}, \textit{etc}. The connection link provides an additional semantic relation between each pair of keypoints and it can be used for many semantic-level tasks. Nevertheless, prevailing keypoint detection methods, \textit{e.g.}\cite{moon2019posefix,sun2019deep}, mainly focus on multi-person human pose estimation, which aims at recognizing and localizing anatomical keypoints (or body joints) and human skeleton connections. On the other hand, existing object (or rigid body) keypoint localization approaches, \textit{e.g.}\cite{pavlakos20176,zhou2018unsupervised,suwajanakorn2018discovery,park2019few}, always fail to successively explore CL among keypoints. Few studies tackled the problem of simultaneously inferring keypoint and encoding their semantic connection information for multi-class instances. 

Inspired by several prevalent bottom-up approaches in the field of multi-person pose estimation, such as \cite{cao2017realtime,kreiss2019pifpaf,papandreou2018personlab}, which directly localize all keypoints from multi-instances and group keypoints into persons to find skeleton connection, we wonder what if to apply a similar mechanism on multi-class object keypoint detection and CL prediction. However, the previous methods cannot be simply grafted onto this issue. The conventional approaches stack each heatmap as a particular class of keypoint for single-type pose estimation, especially for classes of nodes in human pose estimation which belongs to only one category: person. When we move to multi-class rigid bodies, the conventional approaches are inefficient and costly, even impossible, since each category contains numerous classes of keypoints. Consequently, two key factors should be addressed: \textit{(1) how to deal with multi-class instances} and \textit{(2) how to encode the semantic keypoints and their connection links}. Since geometric contextual relations are the keys to identify the keypoint-instance affiliation, these relations spontaneously construct a graph, which consists of nodes (keypoints) and edges (relations between keypoints). A graph structure could be constructed. Besides, the crucial semantic information -- the category of instance that a given keypoint belongs to, should also be considered for CL prediction for multi-class objects. So we can then discover geometrically and semantically consistent keypoints across instances of different categories.

There are two mainstream approaches in instance keypoint detection field, category-specific detection and category-agnostic detection. For category-specific methods like KeypointNet \cite{suwajanakorn2018discovery} in Fig. \ref{fig:approach}(a), they typically group each keypoint as an independent category concerning a given classified target, which is extremely ineffective, costly, and practically inapplicable for building a graph for keypoint and CL prediction for objects with a varying number of parts. For category-agnostic methods like SVK \cite{goel2020shape} and StarMap \cite{zhou2018starmap}, StarMap in Fig. \ref{fig:approach}(b) typically introduces a hybrid representation of all keypoints and compress those belonging to one target as the same class to a single heatmap, and then lift them into 3D space to adjust their locations in 2D. However, StarMap is also costly due to the massive 3D geometry information, such as depth map or multi-view consistency. Besides, the connection information is lost due to the lack of semantic property. Thus, we hope to find a novel, economical, yet powerful approach to directly work in the 2D images.

\begin{figure*}[t]
\centering
\includegraphics[width=15cm]{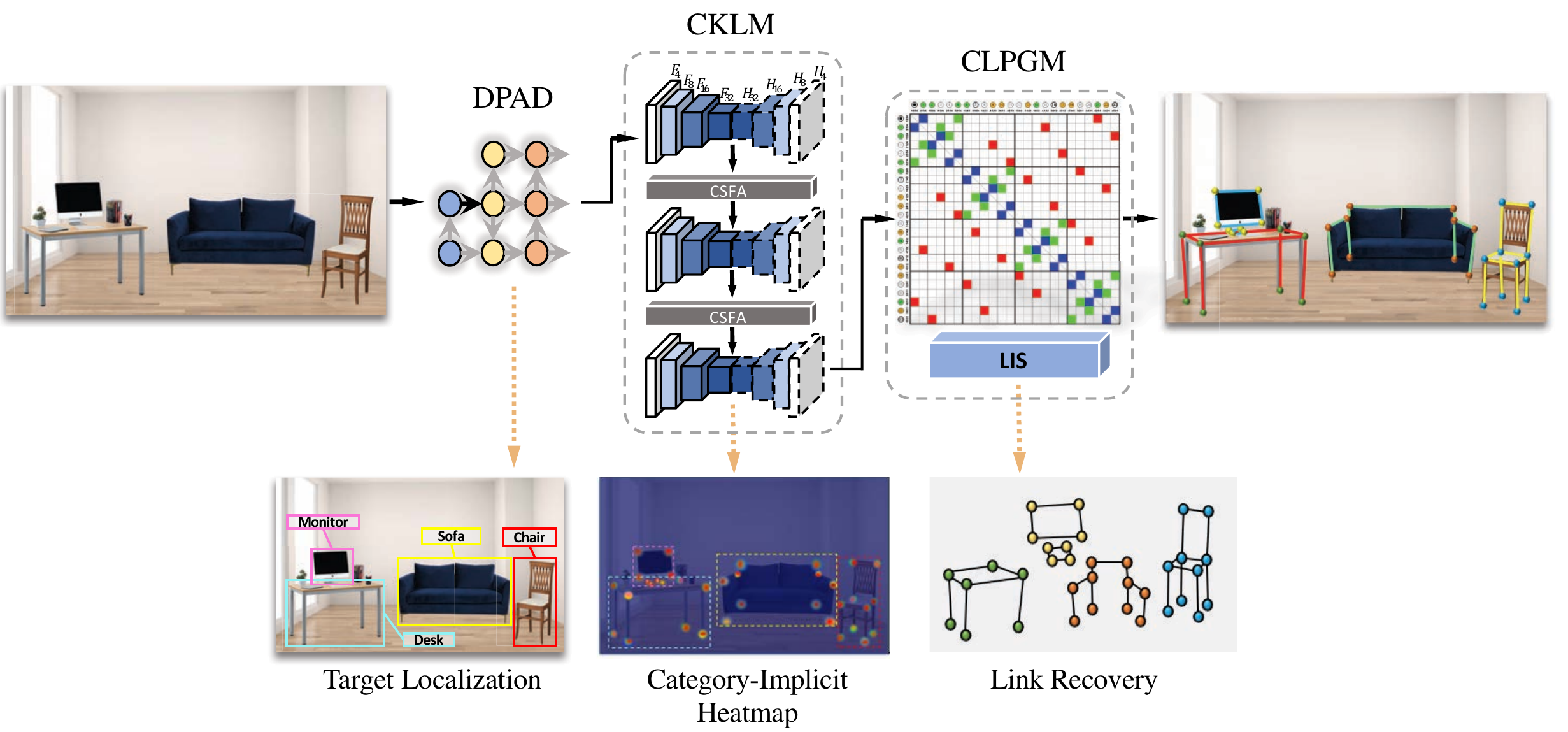}
\caption{Illustration of the proposed KLPNet that is composed of DPAD,  CKLM (detailed in Fig. \ref{fig:cross}) and CLPGM. After objects are allocated by DPAD, category-implicit nodes that belong to the same target are classified as one class on the single heatmap by Cross-stage feature aggregation (CSFA, detailed in Fig. \ref{fig:agg}). $F$ and $H$ are down-sampling and up-sampling feature maps with subscript representing their size. CLPGM works on the nodes with extracted features and corresponding labels to rejuvenate the node links. The Location Instability Stragety (LIS, detailed in Fig. \ref{fig:overlap}) of the inference nodes helps tackle occlusion issues and provide CKLM with feedback to rectify keypoint localization.}
\vspace{-1.2em}
\label{fig:top}
\end{figure*}

To this end, we propose a category-implicit method, Keypoint and Link Prediction Network (KLPNet), as shown in Fig. \ref{fig:top}, including a Deep Path Aggregation Detector (DPAD), a Cross-stage Keypoint Localization Module (CKLM) and a Conditional Link Prediction Graph Module (CLPGM). For the first time, we implement the semantic keypoint detection without converting 2D information to 3D spaces, by virtue of the conditional graph neural network, on rigid bodies. CLPGM recovers the links straightforwardly based on the single heatmap and implicit features of each target extracted from the module DPAD. In CKLM, the Cross-stage feature aggregation scheme is proposed to overcome the ambiguous locations of the category-implicit keypoints on the single heatmap. Specifically, a Location Instability Strategy (LIS) is utilized in GLPGM to disentangle occlusion cases and further respond the defective keypoint localization to the previous module, CKLM. 

The main contributions of this paper can be summarized as follows: (1) To the best of our knowledge, we present the first category-implicit and graph embedding approach, KLPNet, to effectively construe keypoints for instances under multiple categories with a flexible number of semantic labels, and further predict its conditional connection links (2) We propose the keypoint data representation without redundant 3D geometry information, whose location can be adjusted through Cross-stage feature aggregation in a coarse-to-fine manner (3) A novel link prediction module, CLGPM, can enhance the node links based on the single heatmap and extracted implicit features of each target, providing the geometric and semantic information for link recovery. An innovative strategy, LIS in CLGPM, are capable of disentangling the cases with occlusions (4) We explore a deep path aggregation detector to localize the targeting instances precisely.

%-------------------------------------------------------------------------

\section{Related Work}
% \subsection{Instance Detection}
% Deep object detection algorithms can mainly be grouped into two categories: anchor-based and anchor-free ones. The former typically generates bounding boxes with predefined anchors. Meanwhile, anchor-free methods waive the anchor-based region classification and implicit feature learning, but localize bounding boxes according to extraordinary points. YOLO series \cite{redmon2018yolov3,bochkovskiy2020yolov4} learns anchor shape priors. ExtremeNet \cite{zhou2019bottom} detects four extreme points (top-most, left-most, bottom-most, right-most) and one center point of objects using a standard keypoint estimation network.
\subsection{Keypoint Estimation and Geometric Reasoning}
The prominent thoughts of keypoint detection on rigid bodies concentrate on the feature extraction as a two-stage pipeline: identify the localization of each object on the image, and then solve the single object pose estimation problem based on the cropped target. Stacked hourglass  \cite{newell2016stacked} stacks hourglasses that are down-sampled and up-sampled modules with residual connections to enhance the pose estimation performance. Based on the stacked hourglass network, Cascade Pyramid Network\cite{chen2018cascaded} address the pose estimation by adopting two sub-networks: GlobalNet and RefineNet. GlobalNet locates each keypoint to one heatmap that is easier to detect, then RefineNet explicitly address the 'hard' keypoints that requires more context information and processing rather than the nearby appearance feature. The multi-stage pose estimation network (MSPN) \cite{li2019rethinking} extends the GlobalNet to multiple stages for aggregating features across different stages to strengthen the information flow and mitigate the difficulty in training. 

The conventional approaches stack each heatmap as a special class of keypoint for single types, including as left-top, right-top, left-bottom, right-bottom ones. Based on well-defined semantic keypoints, when we move to keypoint detection on multi-class objects, such as bus, chair, ship etc, it is ineffective and costly to train $N\times C$ classes keypoints, where $N$ represents the total number of keypoints of each category, and $C$ is the number of categories. In addition, the value of $N$ varies in different categories. In terms of merging keypoints from multiple targets, consistent correspondences should be established between different keypoints across multiple target categories, which is difficult or sometimes impossible. Besides, category-specific keypoint encoding fail to capture both the intra-category part variations and the inter-category part similarities.

To solve the above-mentioned issues, category-agnostic approaches project the keypoints that belong to the same target to the same category on one heatmap, and then provide additional information to convert 2D image to 3D space for pose estimation. StarMap \cite{zhou2018starmap} mixes all types of keypoints using a single heatmap for general keypoint detection. KeyPointNet \cite{suwajanakorn2018discovery} considers the relative pose estimation loss to penalize the angular difference between the ground truth rotation and the predicted rotation using orthogonal procrustes. Both approaches have to convert the 2D to 3D space first, and then adjust the keypoint location by different predefined 3D models. Our approach generate one type of such general implicit keypoints with more explicit geometry property and top-class label. Besides, the geometric adjustment works in multiple stages in 2D space, increasing the cost-efficiency. Therefore, we consider a novel approach to skip 3D estimation and localize the keypoints more accurately.
\subsection{Graph Link Prediction}
There is a growing interest in the Graph Neural Network (GNN) because of its flexible utilization with body joint relations. \cite{kipf2016variational} introduced the variational graph autoencoder (VGAE) for unsupervised learning on graph-structured data. \cite{li2020r} proposed a model called the Relational-variational Graph AutoEncoder (RVGAE) to predict concept relations within a graph consisting of concept and resource nodes. We propose a new graph structure, CLPGM, to predict the connection among keypoints with different object labels.

\section{Methodology}

\subsection{Deep Path Aggregation Detector}
Inspired by PANet \cite{liu2018path}, we propose DPAD, a Deep Path Aggregation Detector to enhance the localization capability of the entire feature hierarchy by propagating strong responses of low-level patterns. We refer the readers
to \textbf{supplementary material} for its architecture. ResNeXt \cite{xie2017aggregated} is used as the backbone to generate different levels of feature maps $C3 \sim C7$. In addition to these generated feature maps from FPN \cite{lin2017feature}, $C$8 and $C$9, two higher-level feature maps, are created by down-sampling from $C7$. The augmented path starts from the lowest level and gradually approaches to the top. From $C$3 to $C$9, the feature map is down-sampled with factor 2. $\{N3 \sim N9\}$ denote newly generated feature maps corresponding to $\{C3 \sim C9\}$. Each building block takes a higher-resolution feature map $N_i$ and a coarser map $C_{i+1}$ through lateral connection and generates the new feature map $N_{i+1}$. Furthermore, we adopt CIoU \cite{zheng2020distance} to penalize the union area over the circumscribed rectangle's area in IoU Loss. CIoU can improve the trade-off between speed and accuracy for bounding box (BBox) regression problems, and suppresses redundant BBox to increase the robustness of detector for occlusions.

\subsection{Cross-stage Keypoint Localization}
After object targeting, a CKLM would generate detailed localization of all category-implicit keypoints for each classified candidate.
\subsubsection{Single-stage Mechanism}
\begin{figure}[t]
\centering
\includegraphics[width=8.2cm]{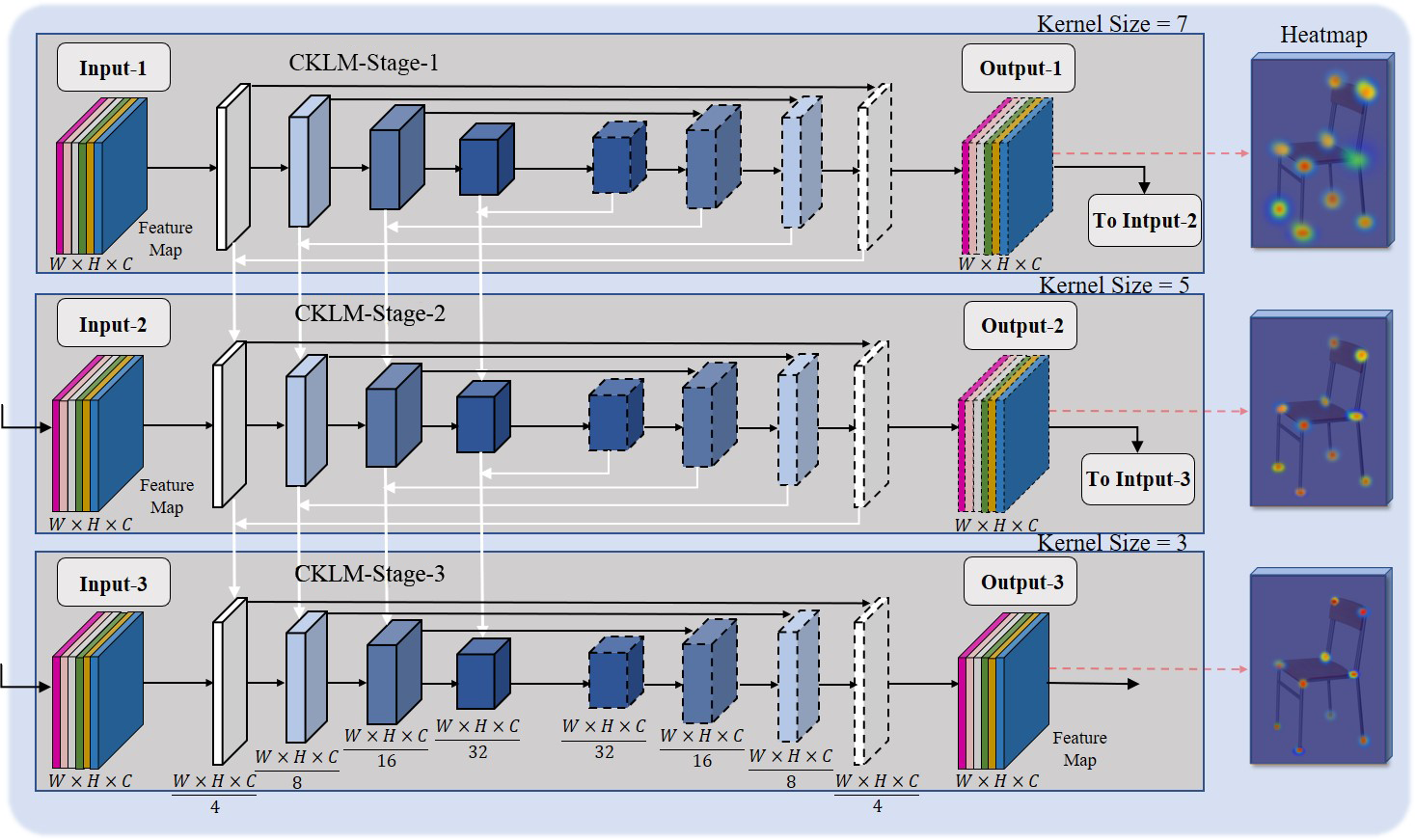}
\caption{An overview of CKLM that consists of three stages with a coarse-to-fine surveillance strategy. A cross-stage aggregation scheme is adopted between adjacent stages (detailed in Fig. \ref{fig:agg}). The coarse-to-fine surveillance strategy utilizes distinctive Gaussian kernel size to boost the keypoint localization performance, as demonstrated on the heatmaps. $W,H,C$ denote the height, weight and channel size for features respectively.}
\vspace{-1.2em}
\label{fig:cross}
\end{figure}

The backbone of the single-stage mechanism is ResNext-101 \cite{xie2017aggregated}. The shallow features have a high spatial resolution for localization but low semantic information for recognition. On the other hand, deep feature layers have rich semantic information but lower spatial resolution due to the convolution and pooling layers. As shown in Fig. \ref{fig:cross}, both spatial resolution and semantic features from distinctive layers are integrated to avoid the unconscious information.

Since we hope to allocate all keypoints from the same target to a single heatmap, they are compelled to set as one class. This single-channel heatmap encodes the image locations of the underlying points. It is motivated by using one heatmap to encode occurrences of one keypoint on multiple persons. The keypoint heatmaps is combined with the confidence maps $H$ and offset maps $\left \{ O_{x}, O_{y}\right \}$. We adopt the binary cross-entropy loss to learn confidence maps with each targeting category and the Smooth L1 loss to update the offset maps.
\begin{ceqn}
\begin{equation}\label{loss_sin}
\mathcal{L}_{kd}=\sum_{i}^{c}\sum_{j}^{k}(\Theta \delta (H-H^{\ast})+\Upsilon  \rho(O_{xy}-O_{xy}^{\ast}) ),
\end{equation}
\end{ceqn}
where $\mathcal{L}_{kd}$ is the keypoint detection loss, $\delta$ is the binary cross-entropy loss, $\rho$ is the Smooth L1 loss, $\Theta$ and $\Upsilon$ indicate the corresponding weights, $c$ is the targeting categories, $k$ denotes the number of keypoints under each targeting category. $H^{\ast}$ and  $O_{xy}^{\ast}$ are the ground truth.
\subsubsection{Cross-stage Feature Aggregation among Multiple Stages}
Feature aggregation could be regarded as an extended residual design in the single-stage mechanism, which is helpful for dealing with the gradient vanishing problem. After the first single-stage module, the single heatmap contains most probable keypoint locations. However, the heatmap from the first single stage is a coarse prediction with abounding noise, even if adequate features have been already extracted in the stage. Even small localization errors would significantly affect the keypoint detection performance. To filter the noise, another two stages are cascaded with the refined surveillance, as shown in Fig. \ref{fig:cross}. Since the Gaussian kernel is used to generate the ground truth heat map for each key point, we decide to utilize distinguishable sizes of kernels, 7, 5, and 3, in these three stages. This strategy is based on the observation that the estimated heat maps from multi-stages are also in a similar coarse-to-fine manner. 

\begin{figure}[t]
\centering
\includegraphics[width=8cm]{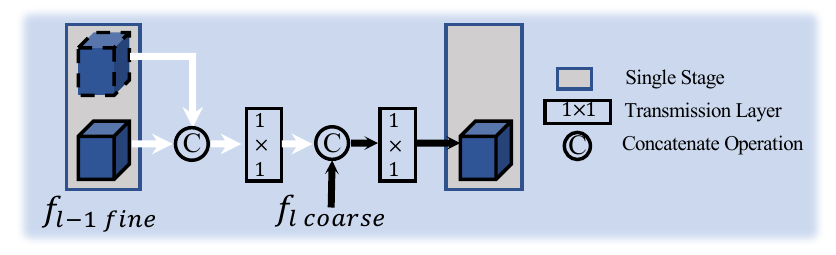}
\caption{Cross-stage feature aggregation scheme. After concatenation, a transmission layer is applied to the features obtained from the previous stage before feature aggregation.}
\label{fig:agg}
\vspace{-1.2em}
\end{figure}

Fig. \ref{fig:agg} shows the cross-stage feature aggregation scheme among multiple stages. $f_{l}$ represents the extracted features in layer $l$. Since we hope to aggregate features from neighbour stages and layers, the coarse-to-fine approach is proposed as follows:
\begin{ceqn}
\begin{equation}\label{addeq_1}
f_{l_{fine}}=T(C(f_{l_{coarse}}, f_{l-1_{out}})),
\end{equation}
\end{ceqn}
\begin{ceqn}
\begin{equation}\label{addeq_2}
f_{l-1_{out}}=T(C(f_{l_{fine\_down}}, f_{l-1_{fine\_up}})),
\end{equation}
\end{ceqn}
where $T$ is the transmission layer which is a $1\times 1$ convolution operation, and $C$ is the concatenation operation. The implementation details of CKLM are reported in \textbf{supplementary material}.

 \subsection{Conditional Link Prediction Graph Module}
Since all the predefined keypoints are localized with agnostic category on the heatmap, we cannot  directly connect them. To solve this issue, a CLPGM is considered to figure out the connection among each key point. In other words, given each target label, we hope to find the adjacency matrix when inferring on an unsupervised learnable graph. Our proposed CLPGM is a pioneer to explore connection links of category-implicit keypoints for multi-class objects.

\subsubsection{Notations} We consider the keypoints under each targeting categories as the node of an undirected, unweighted graph $G = (V, E)$ with $N = |V|$ nodes. Each keypoint is seen as an individual node on gragh $G$. The features of each keypoint extracted from the previous stage are set as the node features and summarized in an $N\times D$ matrix $X_{C}$ with the target label $C$, where $D$ is the degree matrix of the graph. The diagonal elements of the graph's adjacency matrix $A$ are set to 1, which means every node is connected to itself. The adjacency matrix $A$ consists of several $A_{C}$. The stochastic latent variables  $z_{i}$ is summarized in an $N \times F$  matrix $Z$, 
where $N$ is the number of keypoints, $F$ is the depth of features, $Z$ represents the embedding space. 

\subsubsection{Objective Function} CLPGM is built in a top-down manner. Given the training graph, we first disentangle all nodes with $C$ types of labels, where $C$ is the number of target categories. When the training samples from the CKLM arrive, the corresponding nodes with the same target label $C$ are activated and their features $X_{C}$ are updated. After retaining the rest nodes with their features, the whole graph updates the weights and learns the connection. The objective of learning the graph is given as follows:
\begin{ceqn}
\begin{equation}\label{max}
argmax_{\theta_{G}}\prod P(A_{C}|X_{C},\theta_{G}).
\end{equation}
\end{ceqn}
CLPG learns node parameters $\theta_{G}$ that best fits feature maps of training images.

Let us regard $X_{C}$ as a distribution of "node activation entities". We consider the node response of each unit $x \in X_{C}$ as the number of "node activation entities". $ F(X) =\tau \cdot max({f_{x}, 0})$  denotes the number of updated nodes, where $f(X)$ is the normalized response value of $x$ and $\tau$ is a constant parameter.

For a Gaussian mixture model, the distribution of the whole graph at each step is as follows:
\begin{ceqn}
\begin{equation}\label{dis}
P(A_{C}|X_{C},\theta_{G}) = \prod_{X_{C}\in X}q(p_{A_{C}}|X_{C},\theta_{G})^{F(x)},
\end{equation}
\end{ceqn}
where $q(p_{A_{C}}|X_{C},\theta_{G})$ indicates the compatibility of each part of updated sub-adjacency matrix.

Each time $B$ targets are detected from the sample training image, the corresponding sub-adjacency matrices are updated to the whole graph for the advanced training. The features are projected to an embedding space, $Z$, after the first four layers of CLPG. The feature information could be encoded as:
\begin{ceqn}
\begin{equation}\label{cgae}
q\left ( Z|X,A,C \right )=\prod_{i=1}^{N}q\left (z_{i}|X, A, C \right ),
\end{equation}
\end{ceqn}
where $q\left (z_{i}|X, A, C \right )=Gaussian (z_{i}|c_{i},\mu_{i},diag(\sigma _{i}^{2}))$.
In the Equation \ref{cgae}, $\mu$ and $\sigma^{2}$ are the matrices of expected value and variance. Each layer is defined as $tanh(\widetilde{A}XW_{i})$. Denote the target category prediction as $c$, we calculate the piecewise link loss function as follows:
$$ \mathcal{L}_{link}=\left\{
\begin{aligned}
&(c-c_{i})^{2} & &c\neq c_{i} \\
&E_{\varrho }log_{p}[A|Z]-KL[\varrho \left |  \right |p(Z)] & & c = c_{i}\\
\end{aligned}
\right.
$$
where $\varrho=q(Z|X,A)$, $KL[q(\cdot)||p(\cdot)]$ represents the Kullback-Leibler divergence between $q(\cdot)$ and $p(\cdot)$.

\subsubsection{Rejuvenation} For a non-probabilistic variant of the C-GVAE model, we calculate embeddings $Z$ and the rejuvenated adjacency matrix $\hat{A}$ as follows:
\begin{ceqn}
\begin{equation}
\hat{A}=\sigma(ZZ^{T} | C),
\end{equation}
\label{embeddings}
\end{ceqn}
where $\sigma$ is the inner-product operation and $C$ is the targeting category.

Thus, the final loss function of KLPNet can be formulated as:
\begin{ceqn}
\begin{equation}
\mathcal{L}_{KLPNet} = \alpha\mathcal{L}_{kd}+\beta\mathcal{L}_{link},
\end{equation}
\label{loss}
\end{ceqn}
where $\alpha$ and $\beta$ are the predefined constant parameters.

% {\color{red} so what is the value of final $\alpha$ and $\beta$? This part will be discussed in the supplement.}

\subsubsection{Location Instability Strategy of Inference Nodes} 
\begin{figure}[t]
\centering
\includegraphics[width=8cm]{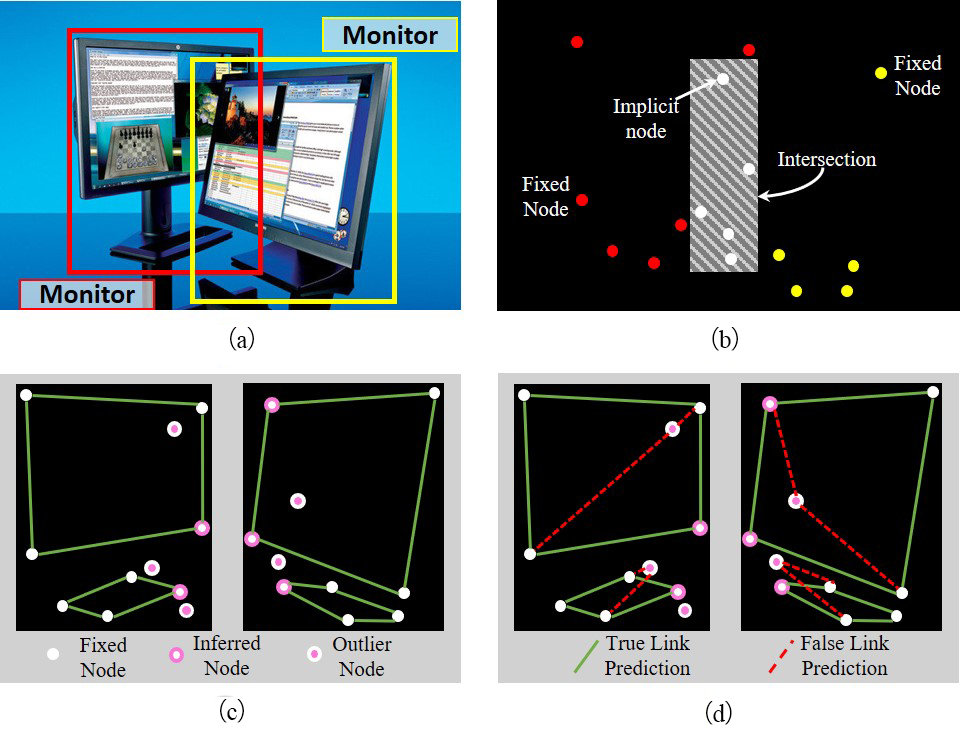}
\caption{Location Instability Strategy. (a) two targets with the same label (monitor) appear in the image, and there exists an occlusion area; (b) the generated heatmap contains three types of nodes (or keypoints): red and yellow nodes are the irrefutable ones which belong to the two targets separately; white nodes are the debatable ones localized in the occlusion area; (c) due to location instability, the debatable ones are distinguished as outliers; (d) illustrations of true (green solid lines) and false (red dash lines) link predictions.}
\vspace{-0.5em}
\label{fig:overlap}
\end{figure}

When multiple targets are occluded, the number of detected nodes may be higher than the predefined number in an area. If these targets belong to distinctive categories, this issue is simplified to generate multi singular heatmaps for each target category. However, if these targets are with the same label, we should design a location instability strategy to infer which nodes are for each overlapped target. As illustrated in Fig. \ref{fig:overlap}(a), an occlusion appears at the center of the image. After keypoint localization through the CKLM, the generated heatmap is illlustrated in the Fig. \ref{fig:overlap}(b). The red and yellow keypoints named fixed nodes are identified to each target. However, the white keypoints with category-implicit nodes on the intersection region (shade part) are confused to be categorized into either of the two monitors to the particular target. 

We first assume that the category-implicit keypoints of the target with the same label share similar features. For example, the similarities of the features belonging to two top-right keypoints of monitors have higher probability of coming from the same instance. Next, We considered that if certain special adjoining nodes always triggered a node, then the inferred node's distance and certain fixed nodes of the object part should not change greatly among targets with the same category label. In the inference part, partial implicit nodes and fixed nodes are utilized to complete the link prediction, where the number of the total nodes should equal to the predefined number of each target category. As shown in the Fig. \ref{fig:overlap}(d), two cases of the predicted results are given, and the incorrect link prediction (red dash line) is deleted. Thus, the implicit nodes are distinguished as inferred node and outlier, as illustrated in Fig. \ref{fig:overlap}(c). 

% \resizebox{10cm}{!}{
% 

\section{Experiments and Results}
\begin{table}[h]\scriptsize
\centering
\caption{Detection performance comparison of different models with backbones including ResNet-101 \cite{he2016deep}, ResNext-101 \cite{xie2017aggregated} and \cite{wang2020cspnet} on MSCOCO Dataset.}
\begin{tabular}{c|ccc|c}
\hline\noalign{\smallskip}
Model & $AP$ & $AP_{50}$ & $AP_{75}$ & Backbone \\
\noalign{\smallskip}\hline\noalign{\smallskip}
Mask R-CNN +FPN \cite{lin2017feature} & 35.7 & 58.0 & 37.8 & ResNet-101 \\
Mask R-CNN +FPN \cite{lin2017feature}& 37.1 & 60.0 & 39.4 & ResNeXt-101 \\
PA-Net \cite{liu2018path} & 42.0 & 65.1 & 45.7 & ResNeXt-101\\
Yolo v4 + PA-Net\cite{bochkovskiy2020yolov4}  & 42.3 & 64.3 & 45.7 & CSPResNeXt50 \\\hline
Ours (DPAD) & \textbf{42.7} & \textbf{64.7} & \textbf{45.8} & ResNeXt-101\\
\noalign{\smallskip}\hline
\end{tabular}
\label{tab:ob}  
\end{table}

\begin{table}[h]\scriptsize
\centering
\caption{Results of the backbone with different layers on MSCOCO Dataset.}
\begin{tabular}{c|ccc}
\hline\noalign{\smallskip}
Method & ResNeXt-50 & ResNeXt-101 & ResNeXt-152 \\
\noalign{\smallskip}\hline\noalign{\smallskip}
AP & 40.6 & 42.7 & 43.3 \\
FLOPs (G) & 4.6 & 7.3 & 12.1 \\
\noalign{\smallskip}\hline
\end{tabular}
\label{tab:ob_base}
\end{table}

% \begin{table}[h]\scriptsize
% \begin{table}[]
% 	\centering
% 	\caption{My table}
% 	\begin{tabular}{|l|l|l|}
% 		\hline
% 		1                       & 2          & 3         \\ \hline
% 		4                       & \multicolumn{2}{l|}{5} \\ \hline
% 		\multicolumn{1}{|l|}{7} & 8          & 9         \\ \hline
% 	\end{tabular}
% \end{table}
%%%%%%%%%%%%%%%
\begin{table*}[h]\scriptsize
\centering
\caption{2D keypoint localization results on Pascal3D+. The results are shown in PCK ($\alpha = 0.1$). Top: our results with the nearest canonical feature as keypoint identification. Bottom: results with oracle keypoint identification. StarMap*: our re-implementation of StarMap \cite{zhou2018starmap} with ResNeXt-101 as the backbone for a fair comparison to ours. Absolute and relative improvements of mean AP (PCK: MCKP; OracleId: 6-DoF SH) are shown in column $\triangle$. CS, CA, CI represents category-specific, category-agnostic, category-implicit methods respectively.}
\begin{tabular}{c|c|cccccccccccc|cc}
% \begin{tabular}{c|llllllllllll|c}
\hline\noalign{\smallskip}
PCK ($\alpha = 0.1$) & TYPE & aero & bike & boat & bottle & bus & car & chair & table & moto & sofa & train & tv & mean &\textbf{$\triangle$}\\
\noalign{\smallskip}\hline\noalign{\smallskip}
MCKP \cite{tulsiani2015viewpoints}& CS & 66.0 & 77.8 & 52.1 & 83.8 & 88.7 & 82.3 & 65.0 & 47.3 & 68.3 & 58.8 & 72.0 & 65.1 & 68.8 &- \\
6-DoF SH \cite{pavlakos20176}& CS & 84.1 & 86.9 & 62.3 & 87.4 & 96.0 & 93.4 & 76.0 & - & - & 78.0 & 58.4 & 84.8 & 82.5&+13.7 \\
StarMap \cite{zhou2018starmap}& CA & 75.2 & 83.2 & 54.8 & 87.0 & 94.4 & 90.0 & 75.4 & 58.0 & 68.8 & 79.8 & 54.0 & 85.8 & 75.5 &+6.6 \\
StarMap* \cite{zhou2018starmap}& CA & 75.4 & 83.8 & 54.2 & 87.8 & 94.1 & 90.5 & 75.9 & 59.1 & 69.2 & 79.7 & 55.8 & 86.8 & 76.0 &+7.2 \\
\hline
Ours (CKLM) & CI& \textbf{85.3} & \textbf{89.1} & \textbf{72.3} & \textbf{93.4} & \textbf{98.5} & \textbf{96.7} & \textbf{82.3} & \textbf{78.4} & \textbf{86.9} & \textbf{89.3} & \textbf{76.1} & \textbf{91.2} & \textbf{86.6} &\textbf{+17.8}
\\\hline\hline
OracleId & TYPE & aero & bike & boat & bottle & bus & car & chair & table & moto & sofa & train & tv & mean\\\hline
6-DoF SH \cite{pavlakos20176} & CA & 92.3 & 93.0 & 79.6 & 89.3 & 97.8 & 96.7 & 83.9 & - & - & 85.1 & 73.3 & 88.5 & 89.0 &-\\
StarMap \cite{zhou2018starmap} & CA& 93.1 & 92.6 & 84.1 & 92.4 & 98.4 & 96.0 & 91.7 & 90.0 & 90.1 & 89.7 & 83.0 & 95.2 & 92.2 &+3.2\\
StarMap* \cite{zhou2018starmap} & CA& 94.2 & 92.9 & 84.8 & 93.1 & 98.8 & 96.5 & 90.4 & 90.6 & 92.1 & 89.8 & 83.3 & 95.7 & 92.6 & +3.7 \\
\hline
Ours (CKLM)& CI  & \textbf{96.2} & \textbf{96.7} & \textbf{89.6} & \textbf{94.8} & \textbf{99.4} & \textbf{98.2} & \textbf{94.3} & \textbf{95.8} & \textbf{96.6} & \textbf{94.2} & \textbf{93.8} & \textbf{98.4} & \textbf{96.4} &\textbf{+7.5}\\
\noalign{\smallskip}\hline
\end{tabular}
\label{tab:pck}  
\end{table*}
\subsection{Setting}
\noindent{{\bf Datasets}}
% \subsubsection{Datasets}
Our models are evaluated on MSCOCO dataset \cite{lin2014microsoft}, Pascal 3D+ \cite{xiang2014beyond} and Object-Net3D \cite{xiang2016objectnet3d}. MSCOCO includes 118k images and 860k annotated objects. We split the train and test dataset as 18K and 2K images, which belongs to 20 types of rigid bodies, such as car, bus, chair, desk etc. Pascal3D+ contains 12 man-made object categories with 2K to 4K images per category. We make use of its category-implicit 2D keypoints for experiments. For fair comparison with state-of-arts, evaluation is
done on the subset of the validation set that is non-truncated and non-occluded. Object-Net3D contains 50k samples, while 20k ones have keypoint annotations. We use 19k images with annotation files as the training set and 10k images for test datasets.\\
\noindent{{\bf Evalutation Metrics}}
% \subsubsection{Evalutation Metrics}
The main metric is Average Precision (AP) over a dense set of fixed recall threshold. We show AP at Intersection-over-Union (IoU) or Object Keypoint Similarity (OKS) threshold $50\%$ ($AP_{50}$), $75\% $($AP_{75}$), and averaged over all thresholds between 0.5 and 1. For keypoint localization, two protocols, Percentage of Correct Keypoints (PCK) and Oracle assigned Keypoint Identification (OracleId), are  considered to evaluate the performance of the models. Floating Point Operations per Second (FLOPs) is used to measure computational performance.\\
\noindent{{\bf Implementation Details}}
% \subsubsection{Implementation Details} 
The KLPNet is implemented in the PyTorch framework and trained on four Nvidia RTX 2080 Ti GPU in 180k iterations. We adopted Mosaic to combines four training images with different scales to one image. In addition, batch normalization calculates activation statistics from four different images on each layer. This significantly reduces the need for a large mini-batch size.

\subsection{Analyses}
\noindent{{\bf Choice of Backbone for Detector}}
Since detection boxes are critical for top-down approaches in object estimation, we compare our model with the other four models who has similar backbone, as demonstrated in Tab. \ref{tab:ob}. Based on ResNeXt-101, the corresponding APs of Mask R-CNN + FPN, PANet, and our model are 37.1$\%$, 42.0$\%$, and 42.7$\%$,. Even comparing with Yolov4 + PAN whose backbone is CSPResNeXt50, the performance of our model is slightly better, since our layer is deeper. Tab. \ref{tab:ob_base} demonstrates the AP and FLOPs of our backbone with different layers. The performance gets quickly saturated with the growth of backbone capacity. ResNeXt-101 outperforms ResNeXt-50 by 2.1$\%$ AP and consume 2.7G FLOPs more, but there is only 0.6$\%$ gain from ResNeXt-101 to ResNeXt-152 at the cost of additional 4.8G FLOPs. It is not effective to adopt ResNeXt-152 or larger backbones for a single-stage network.

\begin{table}[h]\scriptsize
\centering
\caption {$i$ -s represents $i$ stages. Top: Performance of different models with single stage (including two-stage Hourglass). Bottom: our CKLM with different number of stages.  All experiments are conducted on MSCOCO minival dataset.}
\label{tab:one_stage}  
\begin{tabular}{c|cc}
\hline\noalign{\smallskip}
 & AP & FLOPs (G)\\
\noalign{\smallskip}\hline\noalign{\smallskip}
1-s Hourglass \cite{newell2017associative} & 54.5 & 3.92 \\
2-s Hourglass \cite{newell2017associative}& 66.5 & 6.14 \\
CPN-GlobalNet \cite{chen2018cascaded} & 66.6 & 3.90 \\
1-s MSPN \cite{li2019rethinking} & 71.5 & 4.4 \\
Our CKLM (1-s) & 72.0 & 5.1 \\
\hline
Ours (2-s) & 75.3 & 10.1 \\
Ours (3-s) & 76.1 & 15.3 \\
Ours (4-s) & 76.4 & 20.1 \\
Ours (5-s) & 76.9 & 23 \\
\noalign{\smallskip}\hline
\end{tabular}
\end{table}

% ingle-stage models, Hourglass, CPN-GlobalNet and single-stage MSPN are compared with our single-stage model

\begin{table*}[t!]\scriptsize
\centering
\caption {Comparison of keypoint estimation on ObjectNet3D+. Note: Ours represents KLPNet with only the basic backbone; Ours$\star$ additionally includes the Cross-stage (three stages) feature aggregation scheme; Ours$\dagger$ contains both Cross-stage feature aggregation scheme and Location Instability Strategy (LIS). Definition of Type and StarMap* are informed in Tab. \ref{tab:pck}. Benchmark of column $\triangle$ is StarMap.}
\begin{tabular}{c|c|ccccccccc|cc}
\hline\noalign{\smallskip}
 &TYPE & bed & sofa & bookshelf & chair & monitor & cabinet & microwave & console & guitar & mean &$\triangle$\\
\noalign{\smallskip}\hline\noalign{\smallskip}

StarMap \cite{zhou2018starmap} &CA& 72.0 & 66.3 & 77.8 & 73.4 & 88.6 & 84.1 & 94.3 & 43.1 & 73.3 &-&-\\
StarMap* \cite{zhou2018starmap} &CA& 72.3 & 66.6 & 77.4 & 73.8 & 89.1 & 84.7 & 94.5 & 43.4 & 74.2 &-&-\\
KeypointNet \cite{suwajanakorn2018discovery} &CS & 74.9 & 71.2 & 74.3 & 76.8 & 81.4 & 88.7 & 95.4 & 57.6 & 71.3 &-&-\\
Ours &CI& 69.6 & 68.9 & 71.8 & 74.1 & 79.6 & 84.3 & 91.4 & 63.7 & 71.6&-&-\\
Ours$\star$ &CI& 81.3 & 75.8 & 76.3 & 77.6 & 85.3 & 87.9 & 96.1 & 68.5 & 74.8&-&-\\
Ours$\dagger$ &CI& \textbf{87.4} & \textbf{81.1} & \textbf{83.4} & \textbf{84.8} & \textbf{89.7} & \textbf{90.1 }& \textbf{77.8} & \textbf{71.3} & \textbf{79.9} &-&-\\ \hline\hline
&TYPE & car & bus & aircraft & mirror & piano & helmet & loudspeaker & knife & printer& mean &$\triangle$\\\hline
StarMap \cite{zhou2018starmap} &CA & 58.7 & 74.8 & 63.5 & 67.9 & 57.1 & 69.7 & 56.7 & 18.2 & 63.4&67.0&-\\
StarMap* \cite{zhou2018starmap}&CA & 58.9 & 75.1 & 63.6 & 67.4 & 57.6 & 70.2 & 56.9 & 18.4 & 63.6&68.0&1.0\\
 KeypointNet \cite{suwajanakorn2018discovery} &CS & 69.8 & 81.3 & 71.4 & 69.6 & 64.8 & 72.6 & 69.4 & 58.7 & 72.7&72.4&5.4\\
Ours &CI & 63.1 & 76.9 & 69.0 & 69.1 & 63.5 & 72.1 & 68.8 & 64.1 & 71.3&71.9&4.9\\
Ours$\star$ &CI& 69.7 & 80.1 & 73.3 & 73.5 & 69.8 & 75.3 & 73.9 & 69.3 & 74.8&76.8&9.8\\
Ours$\dagger$ &CI& \textbf{74.5} & \textbf{85.9} & \textbf{78.9} & \textbf{78.7} & \textbf{72.6} & \textbf{79.1} & \textbf{79.1} & \textbf{72.6} & \textbf{79.7}&\textbf{80.3}&\textbf{13.1}\\
\noalign{\smallskip}\hline
\end{tabular}
\label{tab:final}  
\end{table*}

% \begin{figure*}[t!]
% \centering
% \includegraphics[width=15cm]{figures/result2.png}
% \caption{Examples of the experimental results of our final KLPNet. For each panel (a)-(c), the columns from left to right illustrate input image, category-implicit heatmap, and link recovery based on the localized keypoints on the image, respectively. (a) results with respect to a single object; (b) results with respect to multiple single-class objects (from top to bottom: cases for no occlusion, slight occlusion and normal occlusion); (c) results with respect to multiple multi-class objects (from top to bottom: cases for no occlusion and slight occlusion).}
% \vspace{-1.2em}
% % {\color{red}why $\dagger$ in the first sentence?, rfan}
% \label{fig:results}
% \end{figure*} 

\noindent{{\bf Heatmap Manufacturing with CKLM}}
% \subsubsection{Heatmap Manufacturing with CKLM}
Tab. \ref{tab:pck} shows the performance comparison of the keypoint localization and classification among three state-of-the-art approaches. In the top of \ref{tab:pck}, $ PCK, \alpha = 0.1$ analyze a detected joint is acceptable if the distance between the predicted and the ground-truth joint is within the threshold 0.1. The mean PCK of the evaluation of 6-DoF Stacked Hourglass (SH) and StarMap are 82.5$\%$ and 78.6$\%$, and our model achieves 86.6$\%$. Although the performance of our model is better, the increased values of PCK are only 4.1$\%$ and 8.0$\%$. We emphasize that all counterpart methods are category-specific, thus requiring ground truth object category as input while ours is general. Thus, we did another experiment to prove our thought and identify the improvement of models. The bottom of the table  \ref{tab:pck} factors out the mismanagement caused by improper keypoint ID association. It is obvious that the score is improved to 96.4$\%$. This is quite encouraging since our approach is designed to be a general purpose keypoint predictor on rigid body, indicating that it is advantageous to train a unified network to predict keypoint locations, as this allows to train a single network with more relevant training data.

\noindent{{\bf Single Stage Mechanism for CKLM}}
In this part, we demonstrate the effectiveness of CKLM based on the proposed single-stage mechanism. According to Tab. \ref{tab:one_stage}, the performance of our single-stage model with 72.0$\%$ AP on MSCOCO minival dataset demonstrates the superiority over others. Note that the localization of the keypoints are only considered since the points has no semantic information at this part.

\noindent{{\bf Cross-stage with Feature Aggregation Scheme for CKLM}}
From Tab. \ref{tab:one_stage}, the performance of single-stage Hourglass is inferior. Adding one more stage introduces a large AP margin. Thus, multi-stages of CKLM are further investigated. Tab. \ref{tab:one_stage} shows the performance of divergent stages based on our single-stage model. By comparing with single stage, 2-stage CKLM further leads to a 2.3$\%$ improvement. Introducing the third, fourth and fifth stage maintains a tremendous upward trend and eventually brings an impressive performance boost of 0.8, 0.3 and 0.5 AP improvement on the previous stage. Meanwhile, the cost of three stages is more than two stages 5.2 FLOPs, but an additional 4.8 FLOPs are consumed if the fourth stage is adopted. Finally, the three-stage CKLM is considered to balance the precision and the cost. These experiments indicate that our model successfully pushes the upper bound of existing single-stage and multi-stage CKLM. It obtains noticeable performance gain with more network capacity.

% \begin{figure*}[t]
% \centering
% \includegraphics[width=16cm]{figures/results.jpg}
% \caption{The final results. Left: input image; Middle: category-agnostic Heatmap; Right: Link recovery based on the localized keypoints on the image.}
% \label{fig:results}
% \end{figure*} 
\noindent{{\bf Performance of the Keypoint and Link Prediction Network}}
We compared the performance among three approaches: StarMap, KeypointNet and our KLPNet on ObjectNet3D+. Tab. \ref{tab:final} illustrates the performance of the approaches mentioned above. Both StarMap and KeypointNet convert the image to the 3D spaces with different additional information, which is inefficient and costly. We did the experiment on KLPNet in three different views: the KLPNet with only backbone, KLPNet$\star$ includes the cross-stage feature aggregation scheme; and KLPNet$\dagger$ contains both cross-stage feature aggregation scheme and location instability strategy. From Tab. \ref{tab:final}, KLPNet$\dagger$ achieves the best performance on distinctive categories.
\begin{figure*}[t!]
\centering
\includegraphics[width=14cm]{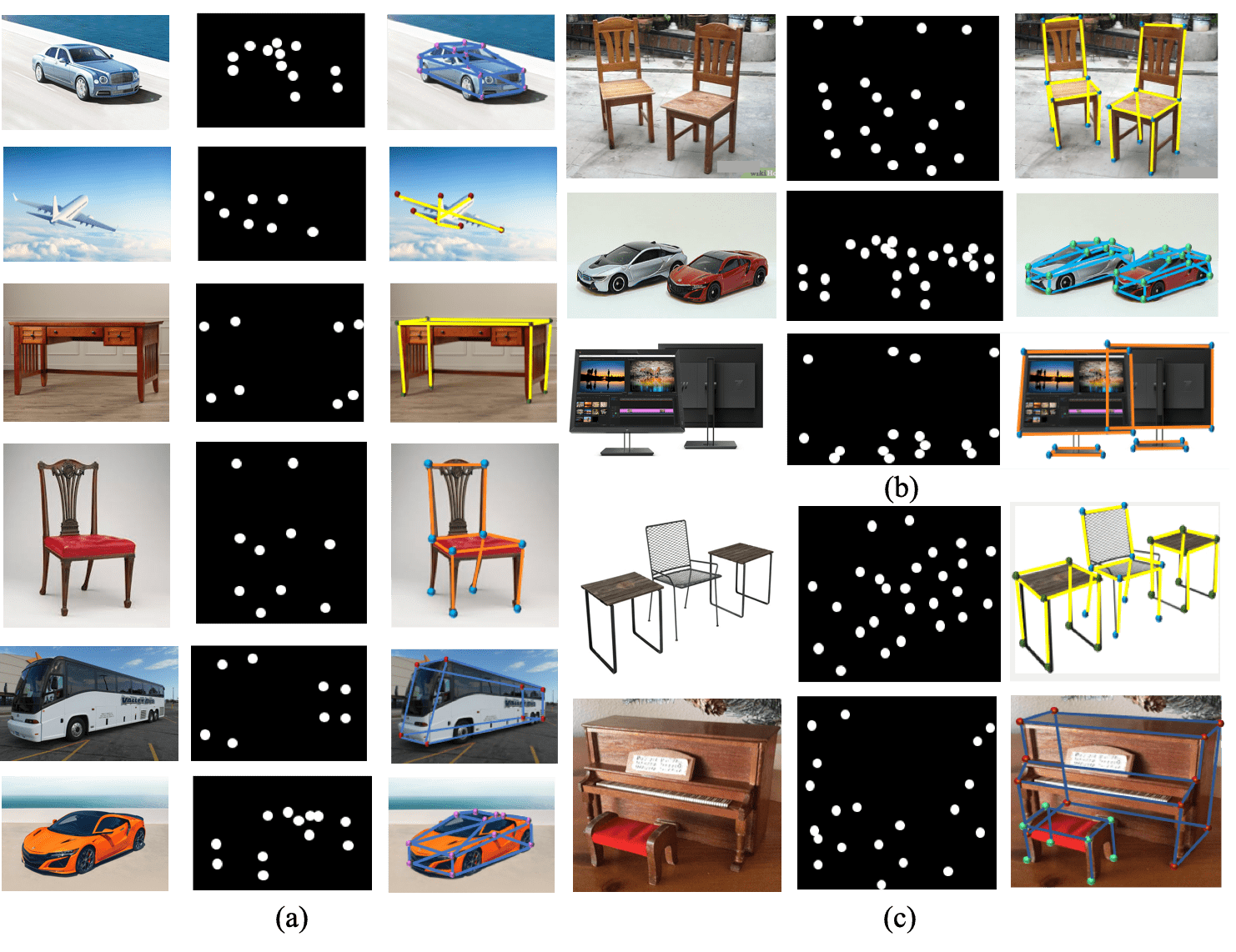}
\caption{Examples of the experimental results of our final KLPNet. For each panel (a)-(c), the columns from left to right illustrate input image, category-implicit heatmap, and link recovery based on the localized keypoints on the image, respectively. (a) results with respect to a single object; (b) results with respect to multiple single-class objects (from top to bottom: cases for no occlusion, slight occlusion and normal occlusion); (c) results with respect to multiple multi-class objects (from top to bottom: cases for no occlusion and slight occlusion).}
\vspace{-1.2em}
% {\color{red}why $\dagger$ in the first sentence?, rfan}
\label{fig:results}
\end{figure*} 
\\
\noindent{{\bf Visual Results of Keypoints and Connection Links}}
Since our approach is the forerunner for link prediction on multi-class rigid bodies, it is hard to compare the quantitative and qualitative results with others. Here we visualize the conditional connection link to illustrate the qualitative performance. According to Fig. \ref{fig:results}, our KLPNet$\dagger$ provides robust connection links in various cases. The semantic information well manifests themselves.

\section {Conclusion}
In this work, we proposed a novel and effective 
simultaneous multi-class object keypoint and connection link rejuvenation approach. The major contributions of this network include: 1) DPAD, a detector capable of localizing category-implicit keypoints accurately; 2) CLPGM, a novel link prediction module capable of recovering the node links straightforwardly based on a single heatmap and the implicit features of each target extracted by DPAP; 3) LIS, an innovative strategy capable of handling occlusion issues; 4) KLPNet, the first end-to-end category-implicit keypoint and link prediction network. The conducted extensive experiments demonstrated both the robustness of our proposed KLPNet, proving the effectiveness of our proposed multi-stage architecture, and meanwhile, showing the state-of-the-art performance on the three publicly available datasets.

\clearpage

\section*{Supplementary Materials}

\section*{A. Deep Path Aggregation Detector}
\begin{figure}[t]
	\centering
	\includegraphics[width=\linewidth]{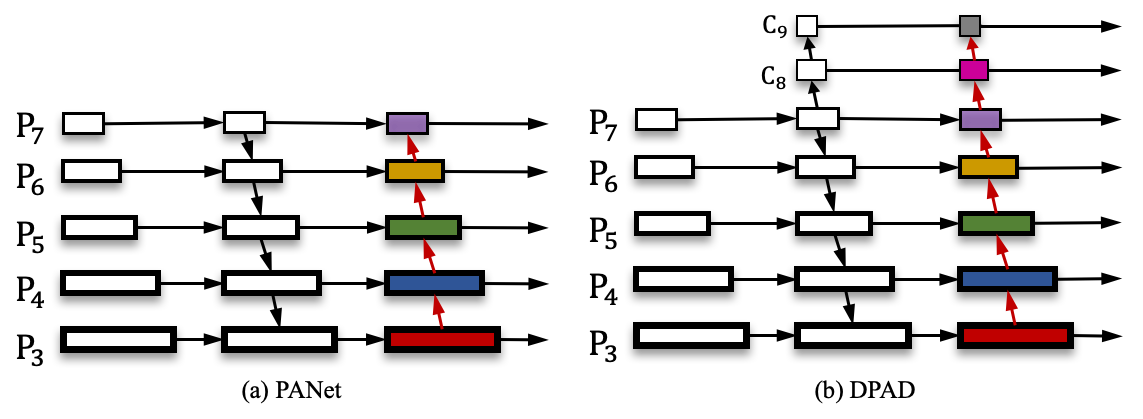}
	\caption{a) The architecture of the PANet. b) The architecture of DPAD. Two higher-level feature maps, C8 and C9, are created by down-sampling from C5. Thus, another two feature maps, N8 and N9, are generated on the neck part. This design enhances the localization capability of the entire feature hierarchy by propagating strong responses of low-level patterns. Additionally, unlike the backbone ResNet utilized in PANet, we use ResNeXt as the backbone in our DPAD.}
	\label{fig:dpa}
\end{figure}

Multi-scale feature fusion aims to aggregate features at different resolutions. Formally, given a multi-scale feature $P_{li}$ at layer $li$, we aim to design an appropriate approach to effectively aggregate different features and update to a deeper layer with renovated features.  The conventional Feature Pyramid Networks (FPN) \cite{lin2017feature} aggregate multi-scale features in a top-down manner, but it is inherently limited by the one-way information flow. Thus, PANet \cite{liu2018path} provides an extra bottom-up path aggregation network. Figure \ref{fig:dpa}(a) illustrates the additional path with red arrows. The neurons in high layers strongly respond to entire objects, while other neurons are more likely to be activated by local texture and patterns. This manifests the necessity of augmenting a top-down path to propagate semantically strong features and enhance all features with reasonable classification capability. The coarser feature map $C_{li}$ at layer $li$ and the generated feature maps $N_{li}$ at layer $li$ with higher resolution can be calculated as:

\begin{ceqn}
	\begin{equation}\label{pan_c}
		C_{li} = Concat(U(C_{l{i+1}}),  g(P_{li})),
	\end{equation}
\end{ceqn}
\begin{ceqn}
	\begin{equation}\label{pan_n}
		N_{li} = Concat(D(N_{l{i-1}}),  g(C_{l{i}})).
	\end{equation}
\end{ceqn}
where $g$ represents the convolutional operations for feature processing, $U$ is usually an upsampling procedure and $D$ is usually a downsampling procedure for resolution alignment, and $Concat$ denotes the concatenate operation.

Inspired by PANet \cite{liu2018path}, we design a deep path aggregation network, DPAD, to enhance the localization capability of the entire feature hierarchy by propagating strong responses of low-level patterns, which is illustrated in Figure \ref{fig:dpa}(b). ResNeXt \cite{xie2017aggregated} is utilized as the backbone network to generate different levels of feature maps, namely $\{C3, C4, C5, C6, C7\}$. Table 1 demonstrates the the superiority of ResNeXt-101 considering both AP and FLOPs. In addition to these feature maps generated from FPN, two higher-level feature maps, C8 and C9, are created by downsampling from C5. The augmented path starts from the lowest level and gradually approaches to the top. From C3 to C9, the spatial size is gradually down-sampled with factor 2. $\{N3, N4, N5, N6, N7, N8, N9\}$ denote newly generated feature maps corresponding to $\{C3, C4, C5, C6, C7, C8, C9\}$. Each building block takes a higher-resolution feature map $N_i$ and a coarser map $C_{i+1}$ through lateral connection to generate a new feature map $N_{i+1}$ as follows:
\begin{ceqn}
	\begin{equation}\label{dpn_c}
		C_{li} = D(C_{li-1})_{i\in ({8, 9} )},
	\end{equation}
\end{ceqn}
\begin{ceqn}
	\begin{equation}\label{dpn_n}
		N_{li} = Concat(D(N_{li-1}),  g(C_{li}))_{i\in ({8, 9} )},
	\end{equation}
\end{ceqn}

\begin{table}[t!]\small
	\centering
	\caption{Detection performance ($\%$) comparison of the ResNeXt models with different layers.}
	\addtolength{\tabcolsep}{-5pt}
	\begin{tabular}{c|ccc|ccc|c}
		\hline\noalign{\small}
		Backbone & $AP$ & $AP_{50}$ & $AP_{75}$ & $AP_{S}$ & $AP_{M}$ & $AP_{L}$ & FLOPs(G) \\
		\noalign{\small}\hline\noalign{\small}
		ResNeXt-50 & 40.6 & 59.3 & 37.2 & 17.8 & 41.6 & 56.4 & 4.6 \\
		ResNeXt-72 & 41.4 & 62.1 & 41.9 & 18.3 & 42.4 & 58.9 & 6.1  \\
		ResNeXt-101 & 42.7 & 64.7 & 45.8 & 20.1 & 45.1 & 64.3 & 7.3 \\
		ResNeXt-124 & 42.9 & 65.1 & 46.3 & 20.8 & 45.4 & 64.9 & 10.6 \\
		ResNeXt-152 & 43.3 & 65.3 & 46.5 & 21.3 & 45.9 & 65.3 & 12.1 \\ 
		\noalign{\small}\hline
	\end{tabular}
	\label{tab:backbone}
\end{table}

\begin{table}[t!]\small
	\centering
	\caption{Detection performance ($\%$) comparison among different models. DPAD$\ast$ has two more higher-level feature maps $C10, C11$, and DPAD$\dagger$ has four more higher-level feature maps $C10, C11, C12, C13$.}
	\addtolength{\tabcolsep}{-5pt}
	\begin{tabular}{c|ccc|cc}
		\hline\noalign{\smallskip}
		Backbone & $AP$ & $AP_{50}$ & $AP_{75}$ & FLOPs(G) & \#Params \\
		\noalign{\smallskip}\hline\noalign{\smallskip}
		PANet           & 42.0 & 65.1 & 45.7 & 7.1 & 78M \\ 
		DPAD$\ast$    & 43.0 & 64.9 & 46.3 & 7.8 & 94.5M\\
		DPAD$\dagger$ & 42.9 & 64.7 & 46.1 & 8.1 & 99.3M\\ 
		\noalign{\smallskip}\hline\noalign{\smallskip}
		NAS-FPN         & 43.1 & 65.3 & 46.5 & 12.1 & 166.5M\\ 
		BiFPN           & 44.4 & 66.4 & 48.3 & 18.7 & 189.2M\\
		\noalign{\smallskip}\hline\noalign{\smallskip}
		DPAD          & 42.7 & 64.7 & 45.8 & 7.3 & 89.8M\\
		\noalign{\smallskip}\hline
	\end{tabular}
	\label{tab:dpa_models}
\end{table}

Unlike PANet \cite{liu2018path}, we remove the mask branch and adopt CIoU \cite{zheng2020distance} to penalize the union area over the circumscribed rectangle’s area in IoU Loss. CIoU can achieve better convergence speed and accuracy for bounding box (BBox) regression problem.
%, and The DIoU NMS \cite{zheng2020distance} developers way of thinking is to add the information of the center point distance to the BBox screening process.

\begin{figure}[t!]
	\centering
	\includegraphics[width=8.2cm]{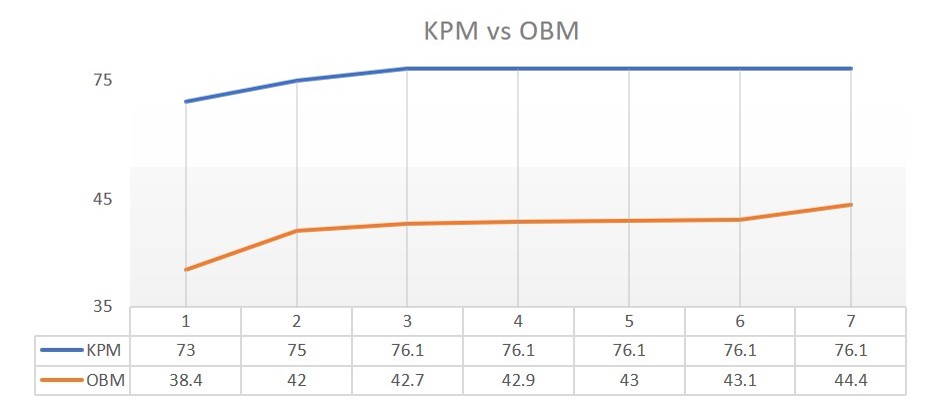}
	\caption{The performance ($\%$) trends of object detection module and keypoint estimation module. In terms of the Keypoint Detection Module (KPM): 1:FPN-based CKLM; 2:PANet-based CKLM; 3.DPAD-based CKLM; 4.DPAD$\ast$-based CKLM; 5. DPAD$\dagger$-based CKLM; 6.NAS-FPN-based CKLM; 7.BIFPN-based CKLM. In terms of the Object Detection Module (OBM):1:FPN; 2:PANet; 3.DPAD; 4.DPAD$\ast$; 5. DPAD$\dagger$; 6.NAS-FPN; 7.BIFPN.}
	\label{fig:kpm_obm}
\end{figure}

We have also tried a deeper DPAD and compared the performance with other approaches. Table \ref{tab:dpa_models} shows the performance of DPAD$\ast$ and DPAD$\dagger$, another two deeper DPADs. Based on DPAD, the former has two more higher-level feature maps $C10$ and  $C11$, and the other has four more higher-level feature maps $C10, C11, C12$, and $C13$. From Table \ref{tab:dpa_models}, we achieve an advance of $0.3\%$ in AP for DPAD$\ast$, but the FLOPs increases by 0.5G. Even the module is designed deeper, like in DPAD$\dagger$, the AP will even decrease, which indicates the precision of the module does not increase w.r.t. the deeper feature layers. Another two approaches, NAS-FPN \cite{ghiasi2019fpn} and BIFPN \cite{tan2020efficientdet} can achieve higher precision, but the FLOPs and the number of parameters are also huge. Thus, we discuss what influences object detector on the top-down keypoint detection in Figure \ref{fig:kpm_obm}. 

As shown in Figure \ref{fig:kpm_obm}, the orange line is the trend of the object detector performance; and the blue one illustrates the performance of the top-down keypoint detection. When the object detector's precision is low, the top-down keypoint detection performance will be more dependent on the objector detector. However, when the object detector's AP is larger than 42.7$\%$, the precision of the top-down keypoint detection becomes saturated; namely, the top-down keypoint detection performance does not heavily rely on the object detector. Thus, a trade-off between the precision and the cost is the key to design an efficient top-down keypoint detector. Based on this prospect, we adopt DPAD as the final object detector in our system.

\begin{table}[t!]\small
	\centering
	\caption{Comparison of a 3-stage CKLM with different corse-to-fine strategies on COCO minival dataset. For each strategy, the number in the table represents the kernel size. The kernel size controls the fineness of supervision and a smaller value indicates a finer setting.}
	\begin{tabular}{c|ccccccc}
		\hline\noalign{\smallskip}
		Setting & 1 & 2 & 3 & 4 & 5 & 6 & 7\\
		\noalign{\smallskip}\hline\noalign{\smallskip}
		Stage 1 & 7 & 5 & 7 & 7 & 7 & 5 & 5\\ 
		Stage 2 & 7 & 5 & 5 & 7 & 5 & 5 & 3\\
		Stage 3 & 7 & 5 & 5 & 5 & 3 & 3 & 3\\
		AP($\%$) & 75.0 & 74.6 & 75.1 & 74.8 & 75.3 & 74.4 & 74.1 \\ 
		\noalign{\smallskip}\hline
	\end{tabular}
	\vspace{-1em}
	\label{tab:kernel_ap}
\end{table}

\begin{table}[t!]\small
	\centering
	\caption{The architecture of the single stage in CKLM.}
	\addtolength{\tabcolsep}{-6.5pt}
	\begin{tabular}{c|cccc}
		\hline\noalign{\smallskip}
		DownSampling Path & Type & Number & Type & Number\\
		\noalign{\smallskip}\hline\noalign{\smallskip}
		DS-1 & BottleNeck-4 & 3 & BottleNeck-3 & 4 \\
		DS-2 & BottleNeck-3 & 5 & BottleNeck-4 & 3 \\
		DS-3 & BottleNeck-4 & 2 & BottleNeck-3 & 5 \\
		DS-4 & BottleNeck-3 & 5 & BottleNeck-4 & 3 \\
		\noalign{\smallskip}\hline\noalign{\smallskip}
		UpSampling Path & Type & Number & Type & Number\\
		\noalign{\smallskip}\hline\noalign{\smallskip}
		US-1 & UpUnit-4 & 3 & UpUnit-3 & 4 \\
		US-2 & UpUnit-3 & 5 & UpUnit-4 & 3 \\
		US-3 & UpUnit-4 & 2 & UpUnit-3 & 5 \\
		US-4 & UpUnit-3 & 5 & UpUnit-4 & 3 \\
		\noalign{\smallskip}\hline\hline\noalign{\smallskip}
		Type & Layer & Kernel & Stride & Padding \\
		\noalign{\smallskip}\hline\noalign{\smallskip}
		& Conv	& $1 \times 1$ & $1 \times 1$ & --\\
		& BN + ReLU & -- & -- & -- \\
		& Conv	& $3 \times 3$ & $1 \times 1$ & $1 \times 1$\\
		& BN + ReLU & -- & -- & -- \\
		BottleNeck-4 & Conv	& $3 \times 3$ & $1 \times 1$ & $1 \times 1$\\
		& BN + ReLU & -- & -- & -- \\
		& Conv	& $1 \times 1$ & $1 \times 1$ & --\\
		& BN + ReLU & -- & -- & -- \\
		\noalign{\smallskip}\hline\noalign{\smallskip}
		& Conv	& $1 \times 1$ & $1 \times 1$ & --\\
		& BN + ReLU & -- & -- & -- \\
		BottleNeck-3 & Conv	& $3 \times 3$ & $1 \times 1$ & $1 \times 1$\\
		& BN + ReLU & -- & -- & -- \\
		& Conv	& $1 \times 1$ & $1 \times 1$ & --\\
		& BN + ReLU & -- & -- & -- \\
		\noalign{\smallskip}\hline\hline\noalign{\smallskip}
		& Conv	& $1 \times 1$ & $1 \times 1$ & --\\
		& BN + ReLU & -- & -- & -- \\
		& Conv	& $3 \times 3$ & $1 \times 1$ & $1 \times 1$\\
		& BN + ReLU & -- & -- & -- \\
		UpUnit-4 & Conv	&  $1 \times 1$ & $1 \times 1$ & --\\
		& BN + ReLU & -- & -- & -- \\
		& Conv	& $3 \times 3$ & $1 \times 1$ & $1 \times 1$\\
		& BN + ReLU & -- & -- & -- \\
		\noalign{\smallskip}\hline\noalign{\smallskip}
		& Conv	& $1 \times 1$ & $1 \times 1$ & --\\
		& BN + ReLU & -- & -- & -- \\
		UpUnit-3 & Conv	& $1 \times 1$ & $1 \times 1$ & --\\
		& BN + ReLU & -- & -- & -- \\
		& Conv	& $3 \times 3$ & $1 \times 1$ & $1 \times 1$\\
		& BN + ReLU & -- & -- & -- \\
		\noalign{\smallskip}\hline\noalign{\smallskip}
	\end{tabular}
	\label{tab:architecture}
\end{table}

\begin{figure*}[t!]
	\centering
	\includegraphics[width=12cm]{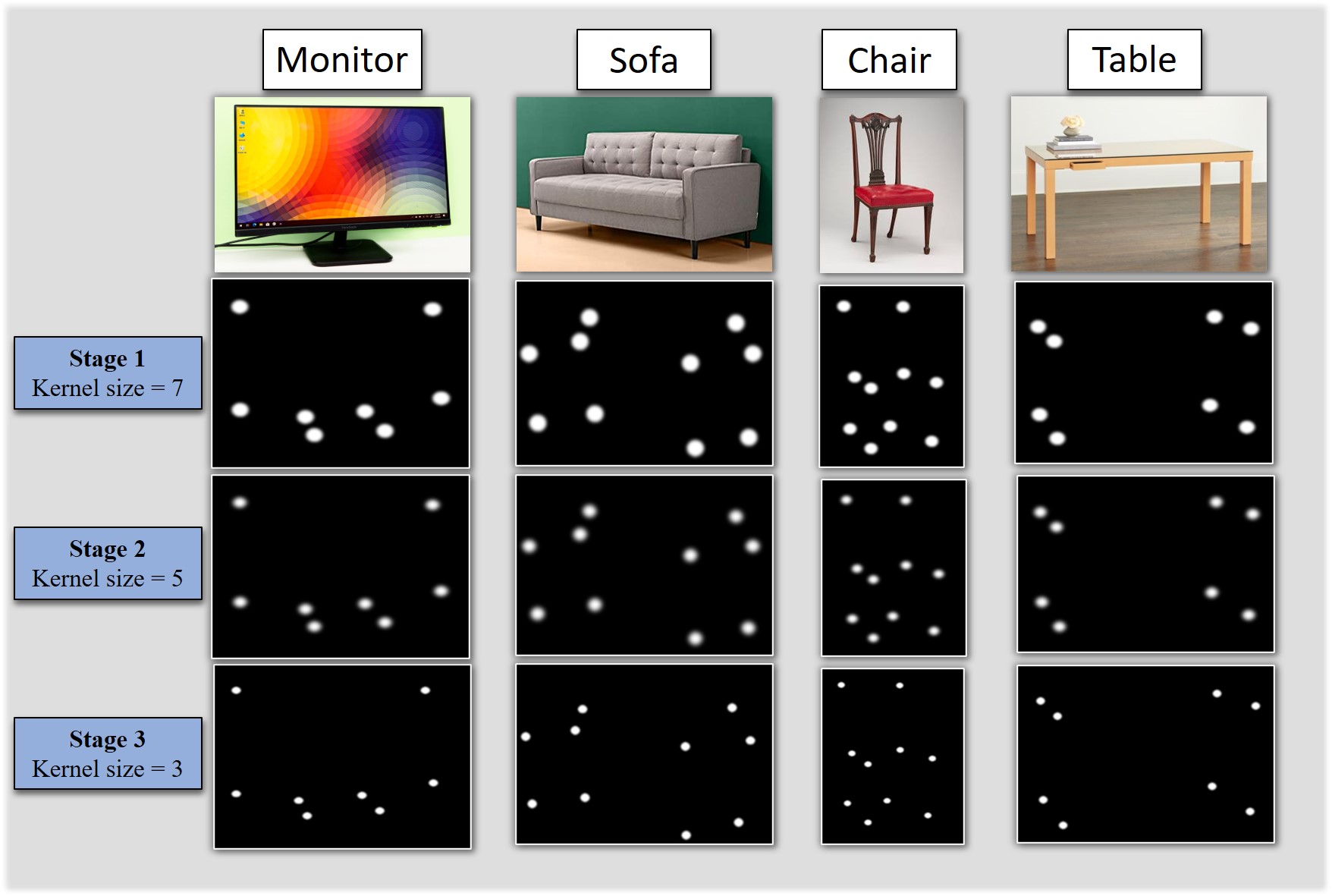}
	\caption{Performance of CKLM with different Gaussian Kernel size.}
	\label{fig:kernel}
\end{figure*} 

\begin{figure}[t!]
	\centering
	\includegraphics[width=\linewidth]{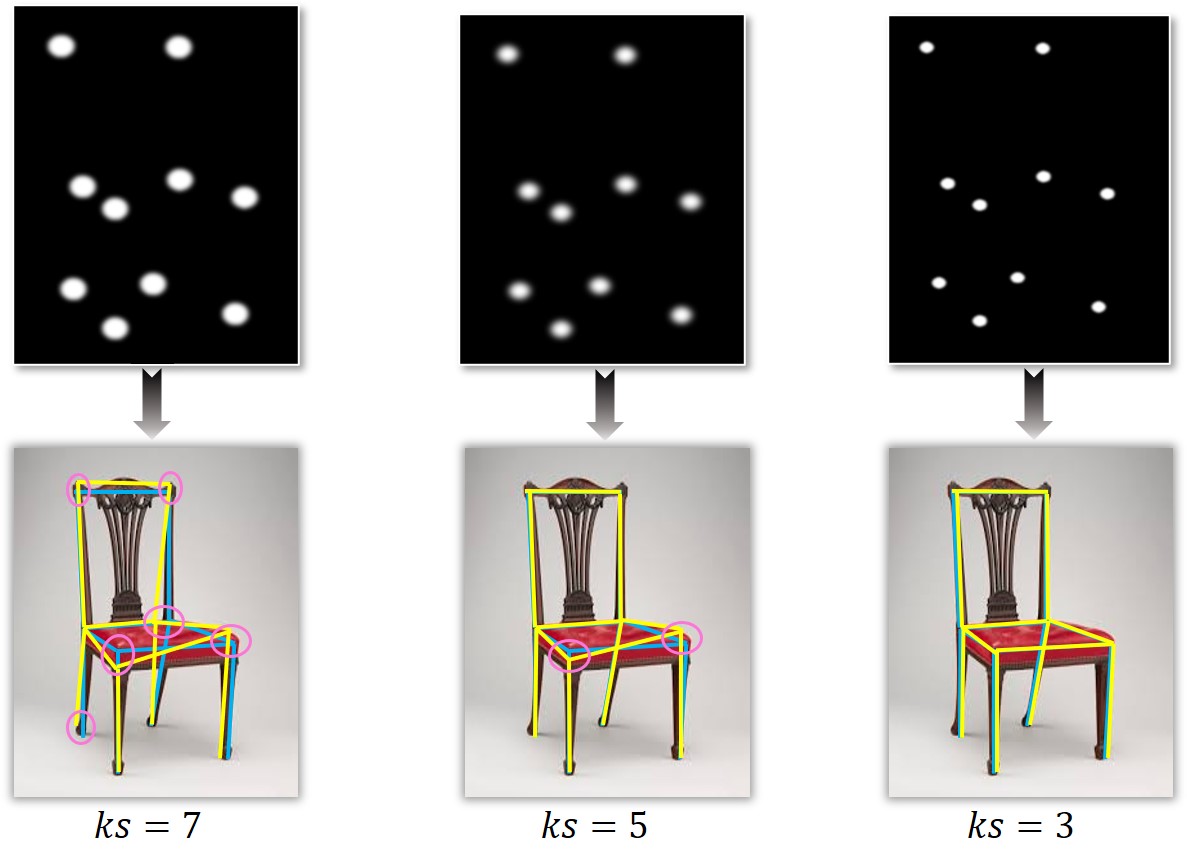}
	\caption{Illustration of a coarse-to-fine strategy. The top row shows the ground-truth heatmaps of the distinctive single stages. From left to right, the Gaussian kernel sizes (ks) of each stage are 7, 5, and 3. The bottom one illustrates the corresponding prediction performance (yellow lines) and the ground-truth annotations (blue lines). The radius of pink circles displays the prediction errors, which are the L2 distances between the ground-truth keypoint position and predicted location.}
	\label{fig:cos_fine}
\end{figure} 

From the top row of Figure \ref{fig:cos_fine}, it is obvious to figure out the differences among the heatmaps with distinguishable sizes of Gaussian kernels, 7, 5, and 3. The strategy is based on the observation that the estimated heatmaps from multi-stages are also in a similar coarse-to-fine manner. The bottom row of Figure \ref{fig:cos_fine} shows an illustrative corresponding predictions (yellow lines) and ground truth annotations (blue lines). The pink circles on both left and center images display the prediction errors, which demonstrates that the proposed strategy is able to refine localization accuracy gradually.

\section*{B. Cross-Stage Keypoint Localization Module (CKLM)}
In the paper, we have discussed the performance of CKLM w.r.t. different number of single stages. The final CKLM module contains three stages to balance the precision and the cost. After a single stage, the single heatmap contains predefined the most probable keypoint locations. However, the heatmap from the first single stage is a coarse prediction with abounding noise, even adequate features have been extracted in the stage. To filter the noise, another two stages are cascaded with a coarse-to-fine surveillance strategy to boost the keypoint localization performance. Since the Gaussian kernel is used to generate the ground truth heat map for each key point, we decide to use distinguishable sizes of kernels, 7, 5, and 3, in these three stages. The distribution of the heatmaps with distinguishable sizes of kernels are demonstrated in Figure \ref{fig:kernel}. 
%Even small localization errors would significantly affect the keypoint detection performance

\begin{table}[t!]\small
	\centering
	\caption {Comparison of Keypoint Estimation Results ($\%$) on ObjectNet3D+. Note: KLPNet represents KLPNet with only the basic backbone; KLPNet$\star$ additionally includes the cross-stage (three stages) feature aggregation scheme; KLPNet$\dagger$ contains both cross-stage feature aggregation scheme and Location Instability Strategy (LIS). LIS of CLPGM provides feedback to CKLM to adjust and fine-tweak the keypoint localization.}
	\addtolength{\tabcolsep}{-8pt}
	\begin{tabular}{c|ccccc}
		\hline\noalign{\smallskip}
		Method & bed & sofa & bookshelf & chair & monitor\\
		\noalign{\smallskip}\hline\noalign{\smallskip}
		KLPNet  & 69.6 & 68.9 & 71.8 & 74.1 & 79.6\\
		KLPNet$\star\ $  & 81.3 & 75.8 & 76.3 & 77.6 & 85.3\\
		KLPNet$\dagger\ $  & 87.4 & 81.1 & 83.4 & 84.8 & 89.7\\
		\noalign{\smallskip}\hline\hline\noalign{\smallskip}
		Method & car & bus & aircraft & mirror & piano\\
		\noalign{\smallskip}\hline\noalign{\smallskip}
		KLPNet & 63.1 & 76.9 & 69.0 & 69.1 & 63.5\\
		KLPNet$\star\ $  & 69.7 & 80.1 & 73.3 & 73.5 & 69.8\\
		KLPNet$\dagger\ $  & 74.5 & 85.9 & 78.9 & 78.7 & 72.6\\
		\noalign{\smallskip}\hline\hline\noalign{\smallskip}
		Method & laptop & diningtable & basket & eraser & flashlight\\
		\noalign{\smallskip}\hline\noalign{\smallskip}
		KLPNet             & 62.9 & 77.4 & 71.2 & 69.3 & 65.3\\
		KLPNet$\star\ $      & 64.3 & 81.3 & 74.3 & 72.8 & 70.8\\
		KLPNet$\dagger\ $    & 71.8 & 84.9 & 78.7 & 76.5 & 74.1\\
		\noalign{\smallskip}\hline\hline\noalign{\smallskip}
		Method & microwave & console & guitar & loudspeaker & knife\\
		\noalign{\smallskip}\hline\noalign{\smallskip}
		KLPNet             & 91.4 & 63.7 & 71.6 & 68.8 & 64.1\\
		KLPNet$\star\ $      & 96.1 & 68.5 & 74.8 & 73.9 & 69.3\\
		KLPNet$\dagger\ $    & 77.8 & 71.3 & 79.9 & 79.1 & 72.6\\
		\noalign{\smallskip}\hline\hline\noalign{\smallskip}
		Method & keyboard & extinguisher & camera & hammer & train\\
		\noalign{\smallskip}\hline\noalign{\smallskip}
		KLPNet             & 71.8 & 65.8 & 70.3 & 64.8 & 73.2\\
		KLPNet$\star\ $      & 74.7 & 69.4 & 75.9 & 69.5 & 76.8\\
		KLPNet$\dagger\ $    & 80.3 & 72.3 & 80.1 & 74.3 & 81.1\\
		\noalign{\smallskip}\hline\hline\noalign{\smallskip}
		Method & cabinet & helmet & coffee maker & printer & blackboard\\
		\noalign{\smallskip}\hline\noalign{\smallskip}
		KLPNet             & 84.3 & 72.1 & 73.4 & 71.3 & 78.5\\
		KLPNet$\star\ $      & 87.9 & 75.3 & 76.8 & 74.8 & 81.3\\
		KLPNet$\dagger\ $    & 90.1 & 79.1 & 81.2 & 79.7 & 86.9\\
		\noalign{\smallskip}\hline
	\end{tabular}
	\label{tab:final}  
\end{table}

Table \ref{tab:kernel_ap} demonstrates the performance of the 3-stage CKLM with disparate Gaussian kernel sizes. We employ distinctive settings for the 3-stage CKLM each time. As shown in Table \ref{tab:kernel_ap}, setting 1 achieves an AP of 75.0$\%$, whose kernel sizes are 7 in all three stages. We tried to degrade the kernel size to 5 in all three stages, however, the AP decreases by 0.4$\%$. It indicates that if we adopt the same kernel sizes, the larger one can present a better performance. We conjecture that this is because the smaller region after the first stage would negatively affect the performance of the detector. Thus, setting 3 and setting 4 are proposed to prove our speculation. When the kernel size in the first stage increases from 5 to 7 in setting 3, the AP is escalated by 0.5$\%$; while the kernel size of the second stage further increases to 7 in the setting 4, the AP is depreciated 74.8 $\%$, even better than the performance in setting 2. It shows that the kernel size in the second stage should be smaller than the one in the first stage. Thus, we decide to diminish the kernel size in the third stage to further validate our hypothesis. In setting 5, the kernel sizes in three stages are set as 7, 5 and 3, respectively. The performance in this setting accomplished the best, and the AP is escalated to 75.3 $\%$. Another two further settings, setting 6 and setting 7, are also made to figure out the performance trends with a smaller kernel size in the first stage. In accordance with expectations, the AP is worse than the one in setting 5. Finally, we adopt the best setting of the kernel sizes: 7, 5 and 3 in our 3-stage CKLM.

Table \ref{tab:architecture} shows the architecture of the single stage of our proposed CKLM. There exist two main branches, the downsampling path and the upsampling path, in each single stage. Each path contains four corresponding layers: DS-1, DS-2, DS-3 and DS-4 for the downsampling path and US-1, US-2, US-3 and US-4 for the upsampling path. The downsampling layer consists of several BottleNeck-4 and bottleneck-3 blocks; while the upsampling layer encompasses several UpUnit-4 and UpUnit-3 blocks. Each BottleNeck and UpUnit block includes distinctive number of convolution layers, batch normalization \cite{ioffe2015batch}, and ReLU \cite{agarap2018deep} activation functions.

\section*{C. Conditional Link Prediction Graph Module (CLPGM)}
In the paper, we construe the details of CLPGM, which includes the Location Instability Strategy (LIS). The LIS is utilized to disentangle occlusion cases under the same category. When multiple targets are occluded, the number of detected nodes may be higher than the predefined number in the overlapped area. If these targets are with the same label, we should design a LIS to infer which nodes are for each overlapped target. Since the details of LIS are already introduced in the paper, we only demonstrate our approach with more samples. 

\begin{figure*}[t]
	\centering
	\includegraphics[width=15cm]{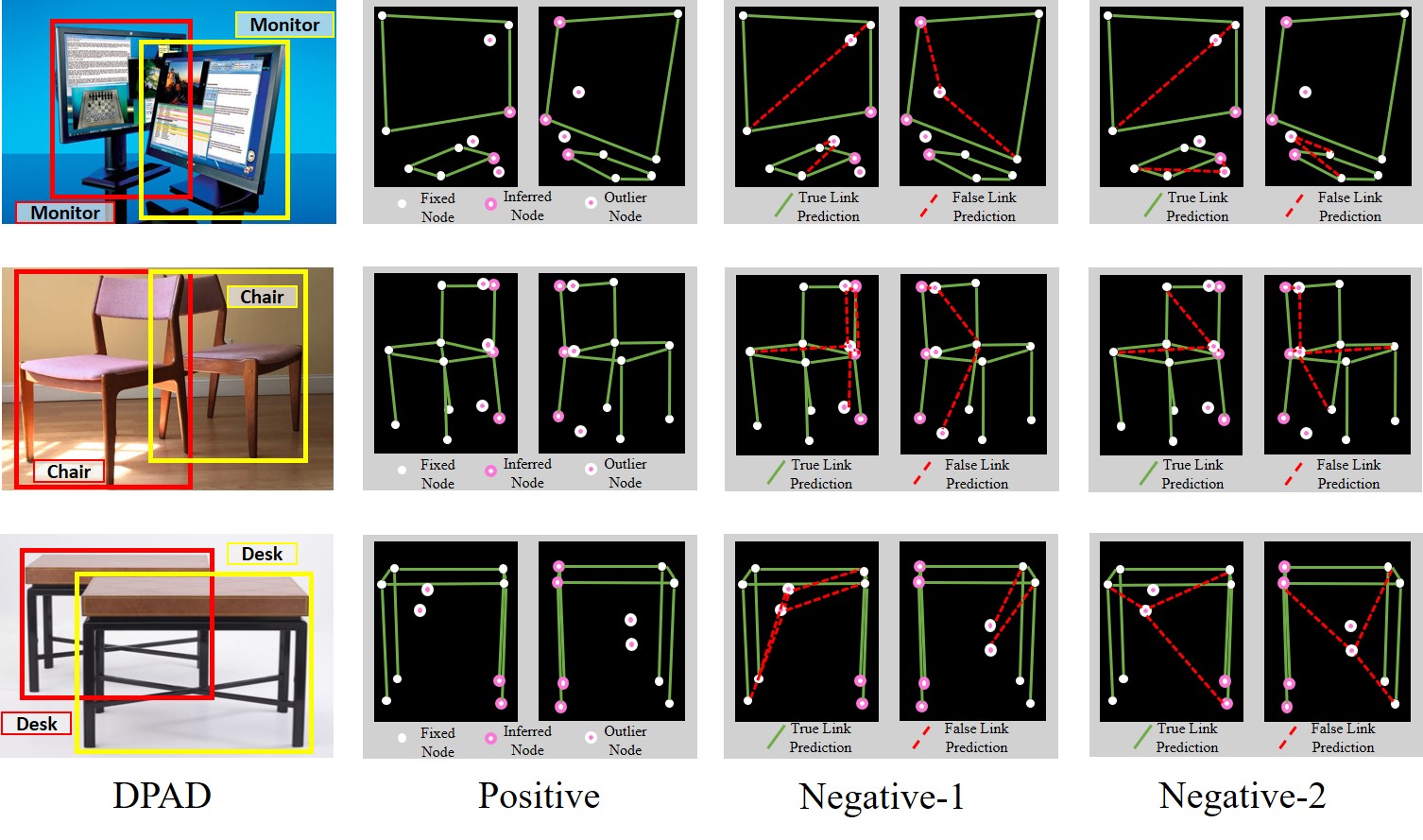}
	\caption{Location Instability Strategy Performance. The left columns indicate the results of the DPAD; The column "Positive" displays the results of LPGM with LIS; The columns "Negative-1" and "Negative-2" illustrate the potential negative samples from the results of the system without LIS.}
	\label{fig:LIS}
\end{figure*} 

Tabel \ref{tab:final} illustrates the comparison among KLPNet, KLPNet$\star$ and KLPNet$\dagger$ on ObjectNet3D+. From Table \ref{tab:final}, KLPNet$\dagger$ achieves the best performance on distinctive categories. Since our approach is the forerunner for link prediction on multi-class rigid bodies, it is hard to compare it with others either quantitatively or qualitatively. Here we visualize the conditional connection link to illustrate the qualitative performance. From Figure \ref{fig:LIS}, our KLPNet$\dagger$ provides correct connection links in various cases and the semantic information well manifests themselves. 
%The images in the left column of the Figure \ref{fig:LIS} are the results of DPAD. The second column displays the results of LPGM with LIS; Meanwhile, the rest two columns are the results of the system without LIS. 

\section*{D. Loss Function}
Recall the total loss Keypoint and Link Prediction Network (KLPNet) is formulated as follows:
\begin{ceqn}
	\begin{equation}
		\mathcal{L}_{KLPNet} = \alpha\mathcal{L}_{kd}+\beta\mathcal{L}_{link},
	\end{equation}
	\label{loss}
\end{ceqn}
where $\alpha$ and $\beta$ are the predefined constant parameters.

\begin{table}[t!]\small
	\centering
	\caption{The performance of the module with different coefficient setting of the loss function}
	\addtolength{\tabcolsep}{-4pt}
	\begin{tabular}{c|ccccccccc}
		\hline\noalign{\smallskip}
		Setting & 1 & 2 & 3 & 4 & 5 & 6 & 7 & 8 & 9\\
		\noalign{\smallskip}\hline\noalign{\smallskip}
		$\alpha$ & 0.1 & 0.2 & 0.3 & 0.4 & 0.5 & 0.6 & 0.7 & 0.8 & 0.9\\ 
		$\beta$  & 0.9 & 0.8 & 0.7 & 0.6 & 0.5 & 0.4 & 0.3 & 0.2 & 0.1\\
		AP & 55.3 & 66.4 & 75.3 & 73.7 & 53.8 & 47.9 & 39.4 & 34.3 & 34.1\\ 
		\noalign{\smallskip}\hline
	\end{tabular}
	\label{tab:loss}
\end{table}

Table \ref{tab:loss} illustrates the performance of the whole module with different settings of the loss. We tried nine settings for the coefficient, $\alpha$ and $\beta$. If the proportion of $\alpha$ is large, the precision of the KLPNet is low; while if it is tiny, the precision is not achievable either. Finally, the setting of $\alpha$ and $\beta$, 0.3 and 0.7, can achieve the best performance. 

\section*{E. Other Applications}
KLPNet can be utilized for keypoint detection on rigid bodies. In this section, we discuss some other applications based on our KLPNet. 

\begin{figure}[t]
	\centering
	\includegraphics[width=\linewidth]{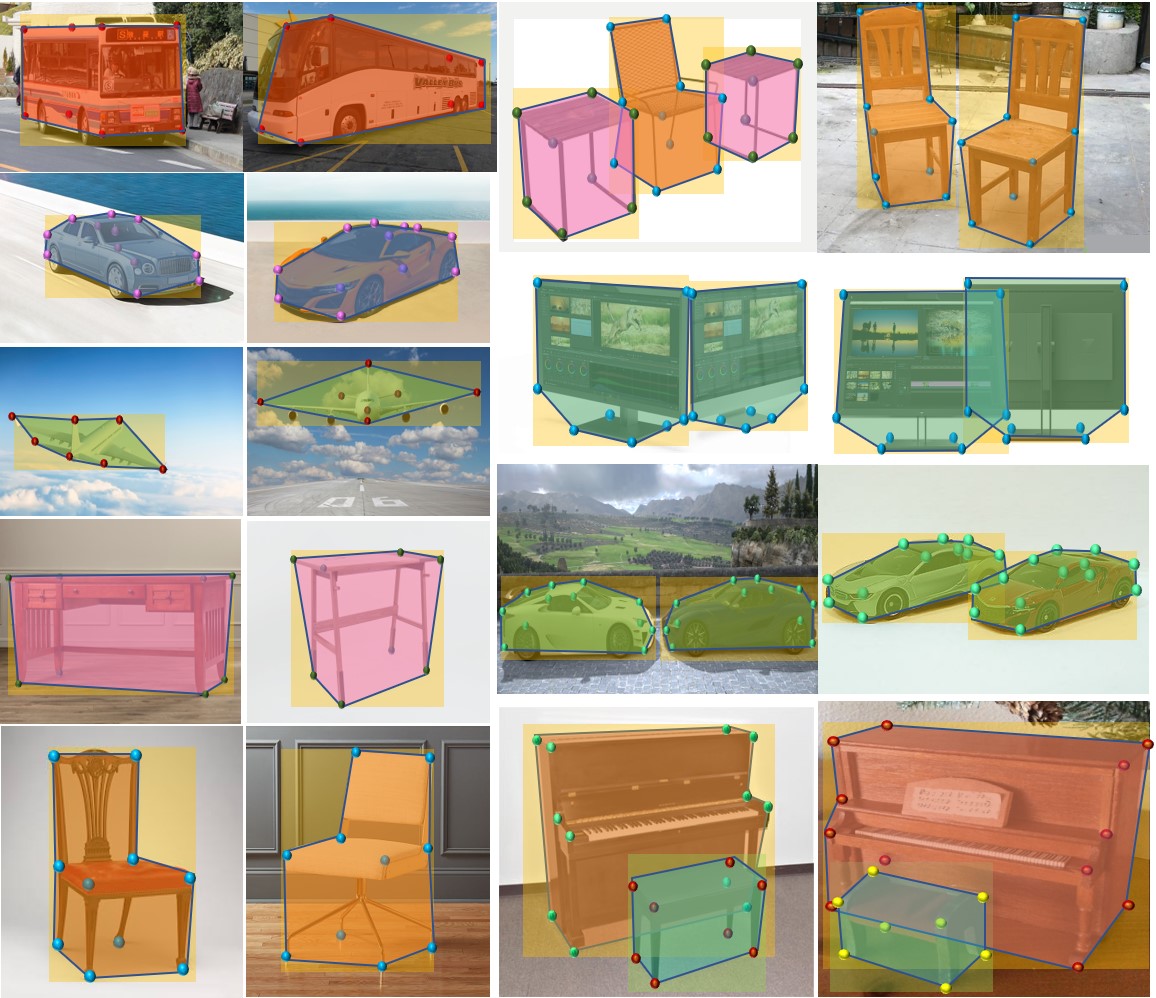}
	\caption{Without CLPGM, the nodes are bounded to get the maximum semantic bounding area. It narrows the interested regions.}
	\label{fig:rod}
\end{figure} 

\subsection*{Refined Object Detection}
If CLPGM is removed from the KLPNet, the category-implicit keypoints are localized in the images. Without CLPGM, we cannot connect the keypoints correctly due to semantic chaos. In this case, we hope to refine the bounding box as a polygon area, which can encircle the target with a narrow yet more accurate area. The results are shown in Figure \ref{fig:rod}. However, we concede this approach is only an attempt to refine the object detection. Some recent object detector \cite{wei2020point} can generate a more accurate and efficient area to encircle the target in the image.

\subsection*{Simultaneous Localization and Mapping (SLAM)}
SLAM is a computational problem of constructing or updating the map of an unknown environment while simultaneously keeping track of an agent's location within it. SLAM \cite{durrant2006simultaneous} \cite{bailey2006simultaneous} contains two subsystems: localization and mapping. Localization is not only the first step but also the key of success to figure out the decision of the whole system. Current approaches \cite{mur2015orb} \cite{cui2019sof} \cite{gomez2019pl} of localization concentrate to pair the landmarks on two frames. We believe that KLPNet offers a new sight to localize the special objects. In terms of SLAM, the details, such as texture, color, are not essential and could be ignored to localize the target. Based on the accurate detection and localization of semantic keypoints in the real world, it is accessible to discern the robot's location and build the real-world mapping. To accomplish it, we need to consider how to get the coordinates in the 3D space, which is discussed in the next part.

\begin{figure}[t]
	\centering
	\includegraphics[width=\linewidth]{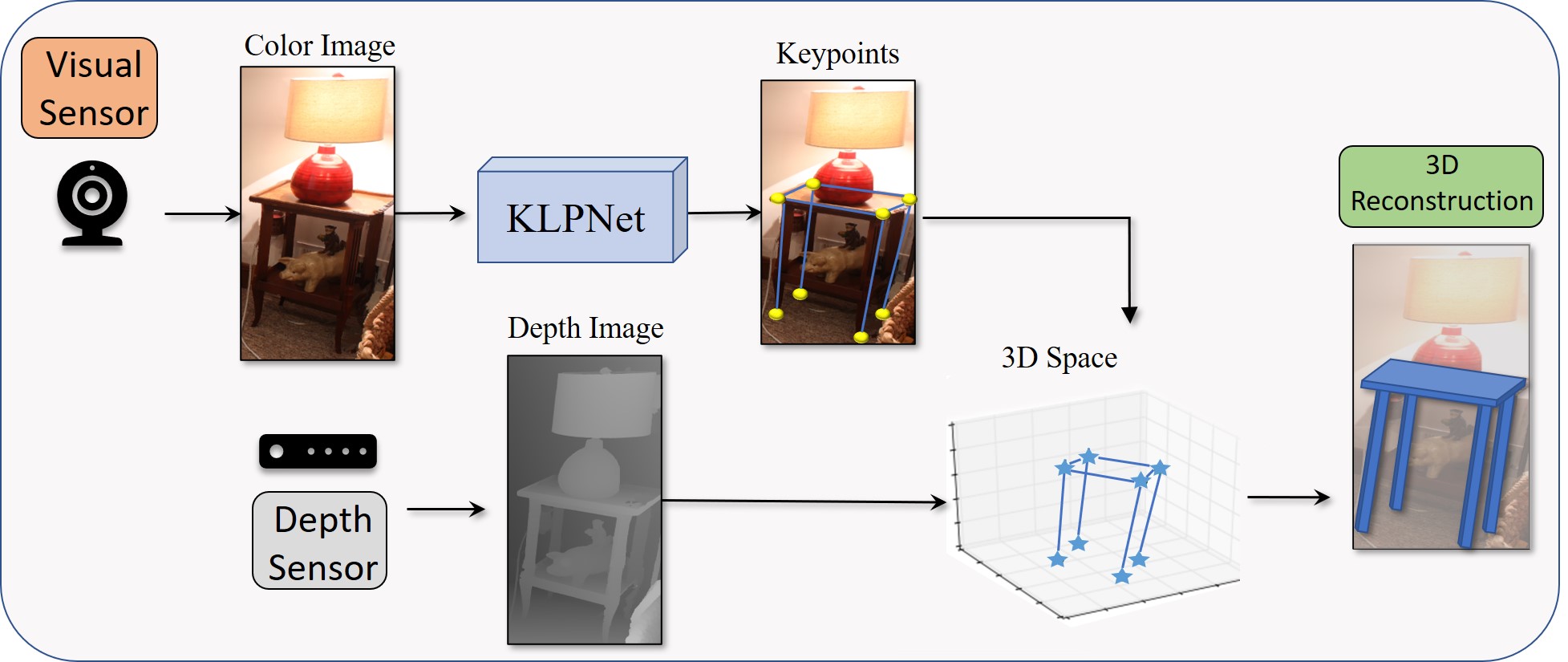}
	\caption{KLPNet with depth map in 3D Applications.}
	\label{fig:keystar}
\end{figure} 

\subsection*{3D Reconstruction and Rendering}
\begin{figure}[t]
	\centering
	\includegraphics[width=\linewidth]{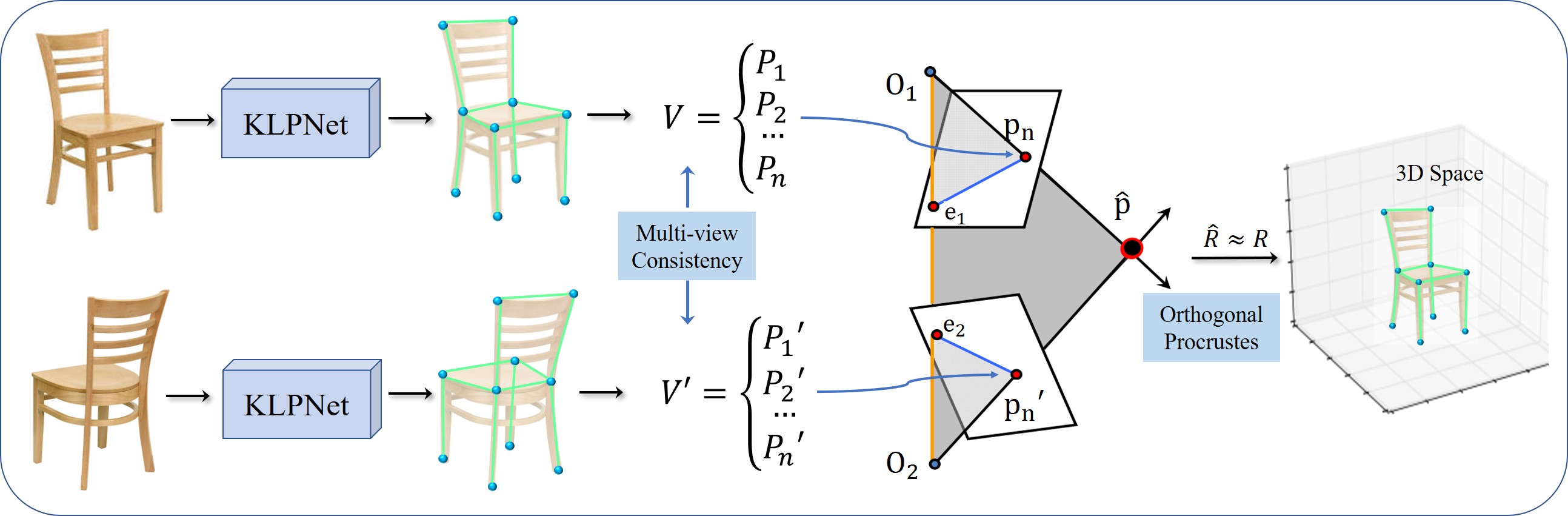}
	\caption{KLPNet with multi-view consistency in 3D Applications. $V$ and $V$' represents two views; $\{P_1,\dotsb,P_i\}$ and $\{P_1$'$,\dotsb,P_i$'$\}$ represent points under the corresponding view; $O$ denotes the original point and $e$ denotes the epipolar.}
	\label{fig:key3d}
\end{figure}

3D Reconstruction \cite{park2020latentfusion} can determine the object's 3D profile, and 3D Rendering \cite{johnson2017mdct} is the final process of creating the actual 2D image or animation from the prepared scene. The first step of both fields is to project the 2D object into a 3D space. Thus, we propose two latent approaches to implement the projection: KLPNet with multi-view consistency and KLPNet with depth map, which are shown in Figure \ref{fig:keystar} and Figure \ref{fig:key3d}.

Figure \ref{fig:keystar} illustrates the first approach for converting 2D targets into a 3D space. Two sensors are utilized to capture useful information, color image and depth map. Using KLPNet, the keypoints can be localized and connected on the 2D heatmap. Depth map affords the space distance of each pixel in the 2D image. After merging the heatmap and depth map, it is conceivable to reconstruct the 3D object in the 3D space.  

The second approach to build the 3D object is to utilize the multi-view consistency instead of the depth map. After obtaining the location and connection of keypoints on the 2D neighbour keyframes, the known rigid rotation ($\textbf{R}$) and translation ($\textbf{T}$) between the two views is provided as a supervisory signal. As shown in the \ref{fig:key3d}, $V_{1}$ and $V_{2}$ are the two views that best match one view to the other. A multi-view consistency loss can be considered to measure the discrepancy between the two sets of keypoints under the ground truth transformation. Once the transformation is corrected, it is conceivable to reconstruct the 3D object in 3D space.

\section*{F. Blemish}

\begin{figure}[t]
	\centering
	\includegraphics[width=\linewidth]{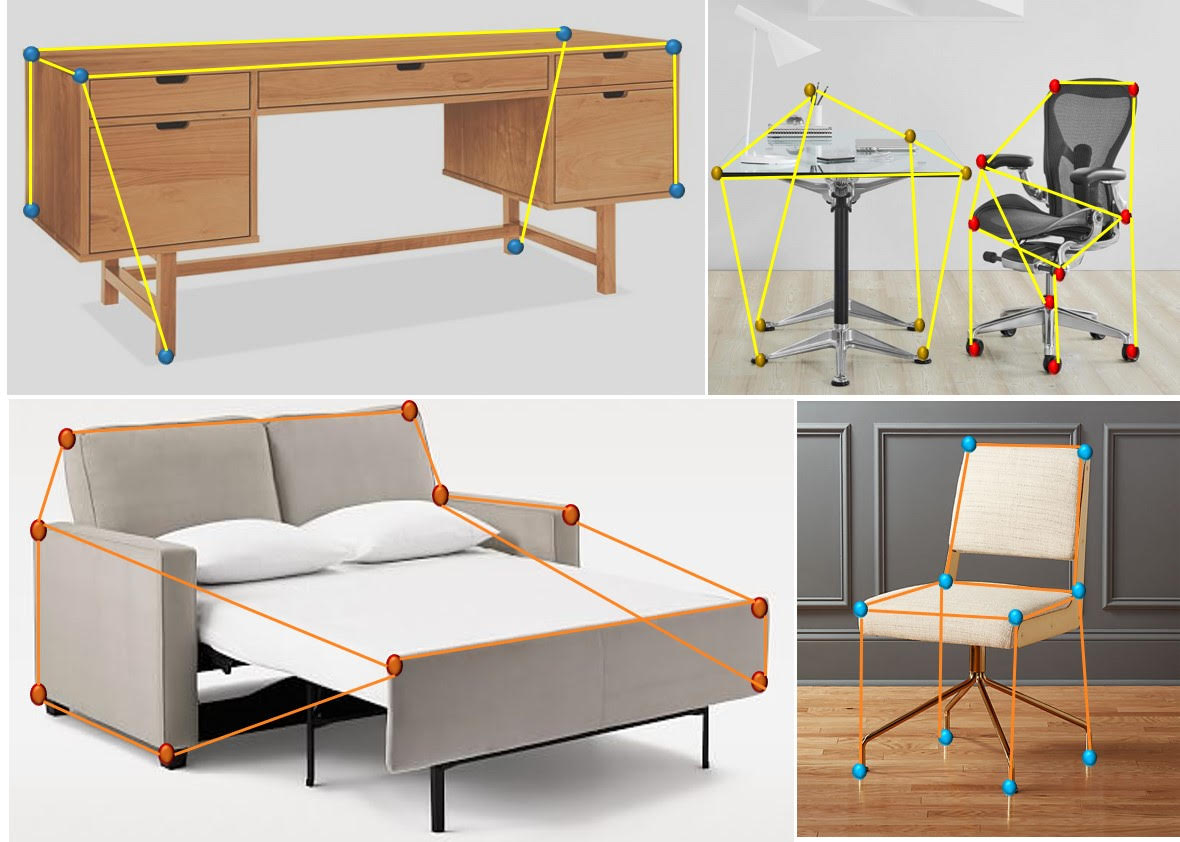}
	\caption{Some unusual cases.}
	\label{fig:bad}
\end{figure} 

During the testing, we also find some blemishes as shown in Figure \ref{fig:bad}. The top-left one is a desk with four additional legs, which lead our model to misjudge the four basic bottom nodes; The bottom-left one is a sofa-bed combination that baffles the model to localize the node accurately since the bed and sofa have a different predefined number of nodes; The top-right contains an unusual desk and a chair who have more legs than normal samples during training; Our model predicts all nodes on the chair correctly but fails to connect all chair legs. We notice that our model cannot self-adapt the node number. The predefined number of nodes per class limits the performance of the model. In the future, a novel supervised approach can be designed to depict more suitable edges for specific geometrical patterns of the objects.

\newpage

{\small
\bibliographystyle{ieee_fullname}

}

\end{document}